# Type-2 Fuzzy Set based Hesitant Fuzzy Linguistic Term Sets for Linguistic Decision Making

Taniya Seth, *Student Member, IEEE*, Pranab K. Muhuri, *Senior Member, IEEE*

*Abstract*—Approaches based on computing with words find good applicability in decision making systems. Predominantly finding their basis in type-1 fuzzy sets, computing with words approaches employ type-1 fuzzy sets as semantics of the linguistic terms. However, type-2 fuzzy sets have been proven to be scientifically more appropriate to represent linguistic information in practical systems. They take into account both the intra-uncertainty as well as the inter-uncertainty in cases where the linguistic information comes from a group of experts. Hence in this paper, we propose to introduce linguistic terms whose semantics are denoted by interval type-2 fuzzy sets within the hesitant fuzzy linguistic term set framework, resulting in type-2 fuzzy sets based hesitant fuzzy linguistic term sets. We also introduce a novel method of computing type-2 fuzzy envelopes out of multiple interval type-2 fuzzy sets with trapezoidal membership functions. Furthermore, the proposed framework with interval type-2 fuzzy sets is applied on a supplier performance evaluation scenario. Since humans are predominantly involved in the entire process of supply chain, their feedback is crucial while deciding many factors. Towards the end of the paper, we compare our presented model with various existing models and demonstrate the advantages of the former.

*Index Terms*— Hesitant fuzzy linguistic term sets, interval type-2 fuzzy sets, multi-criteria group decision making, supplier performance evaluation.

## I. Introduction

DECISION making (DM) is an important task in almost every real life situation. From trivial scenarios like taking a left or right turn to reach a particular place, or to select the best candidate for a vacancy in a company, decisions are crucial, and so is the task of DM. Since most of these tasks involve humans as decision makers (DMR), the involvement of linguistic information in the process of DM is inevitable. This is because it is inherent in humans to express their knowledge linguistically.

The aim of every recent research related to computers, is to make them intelligent enough to emulate the various aspects of a human being. Therefore, for a computer to be able to make decisions like humans, they must first be made to understand the existence of linguistic terms. To enable machines to produce results based on linguistic data, methods of computations on words are required. Supporting this ideology, the computing with words (CWW) paradigm was proposed by Zadeh [1]. The main reasons to employ CWW for linguistic decision making (LDM) problems are as follows [2]:

1) A great part of the human knowledge is defined linguistically.

2) Since precision of words is less than that of numbers, it is appropriate to model words as fuzzy information.

3) There is a cost overhead associated with precision. Hence, the more precise a response is, the more cost it incurs for computation.

The imprecision and vagueness in the expression of human knowledge demands a reliable approach such as the fuzzy linguistic approach (FLA) [3]. The idea behind FLA is to represent the qualitative terms by means of the quintuple linguistic variables to allow computations on them. The quintuple carries information such as the semantics, syntax, name etc. of the associated linguistic variable.

The aforementioned concepts are used till date for developing various LDM approaches to obtain decisions in the form of linguistic terms as outputs, when the inputs provided are also linguistic. However the approaches based on such concepts lack on a serious issue. It is rare for humans to express their response using single words only. They instead tend to use phrases and expressions that contain more than one linguistic term. This is known as the *hesitation* that the DMRs face while presenting their responses. To overcome this restriction in the previously existing approaches, Wang and Hao proposed proportional 2-tuple representation model for linguistic terms [4], wherein it gives the proportional combination of two consecutive linguistic terms. Another linguistic model produces a synthesized term by merging various single linguistic terms [5]. In [6], the authors considered the use of logical connectives and fuzzy relations to obtain results when multiple terms are given.

Even though the aforementioned approaches overcome the restriction of dealing with single linguistic terms as responses, Rodríguez et al. proposed the idea of hesitant fuzzy linguistic term sets (HFLTS) [7], where the responses expected from DMRs are phrases or comparative linguistic expressions (CLE). These expressions are converted into HFLTSs based on a given linguistic term set (LTS). Some recent proposals aim at improving the overall framework of HFLTS such as: [8] wherein weighted HFLTSs have been introduced; distance and similarity measures for HFLTSs [9]; fusing the concept of CLEs with symbolic translation [10], etc. However, each of the existing frameworks of HFLTSs employ only type-1 fuzzy sets (T1 FS) for the representation of linguistic terms. However, T1 FSs are not capable enough to model the linguistic uncertainties [11], hence providing inaccurate and less reliable results.

Based on many recent works, it has been observed that T2 FSs model linguistic uncertainty better. This fact is supported by the statement "*words mean different things to different people*" [12]. To elaborate, a DMR faces two types of uncertainties namely the *intra-uncertainty* and the *inter-uncertainty*, also called the word and expert-level uncertainties respectively. The intra-uncertainty comes into play when the DMR is himself/herself unsure of the exact definition of the linguistic information, hence, varying the endpoints of the intervals which define the linguistic term/information. On the other hand, inter-uncertainty comes into light when there is disagreement amongst various DMRs. To respect every DMRs opinion, the linguistic information must be denoted using at least a T2 FS. Hence, Mendel [13] has rightly said that a linguistic term should be modelled using T2 FSs or higher models to produce scientifically correct representations of linguistic information.

Some linguistic models representing linguistic terms using T2 FSs are [14]-[16]. However, none of these models consider the elicitation of linguistic information from responses that are close to the human cognition. This demands a linguistic model that is robust enough to handle complex responses from DMRs, such as CLEs along with both the word and expert-level uncertainties discussed above.

The framework of HFLTSs were not introduced with T2 FSs until Liu et. al. [17] proposed the idea of incorporating T2 FSs as a representation model for CLEs, called the T2 fuzzy envelope. The authors mentioned that the higher order uncertainty in the CLEs was considered to be obtained from the fuzziness and hesitancy. Even though this idea improves the quality of DM, it still lacks in the context where T2 FSs are most useful.

In [17], semantics of linguistic terms were denoted using T1 FSs. But, as discussed above, the main advantage of introducing T2 FSs within the DM scenarios lies within the idea of handling the expert-level and word level uncertainties. The results obtained by the model in [17] do not take into account the word-level and expert-level uncertainties, therefore leading to results that do not reflect the true thinking of the DMRs.

It must also be pointed out that during the process of obtaining the T2 fuzzy representation for CLEs in [17], fuzzy and hesitant entropies are computed for HFLTSs in order to consider the importance of the linguistic terms in HFLTSs. Given that various entropies are used to evaluate the uncertainties present in the HFLTSs, the process of computing with them is still bound to suffer some information loss. This is because the entropies are computed purely on the indices of the linguistic terms, thus completely ignoring the inherent uncertainty of each linguistic term within the HFLTSs. Also, the method presented in [17], focusses only on triangular fuzzy numbers (TFN), hence neglecting the possibility that the linguistic terms might have a certain range of values where the DMR is certain about the definition of the term.

Keeping in mind all the above points, we propose a novel linguistic model based on T2 fuzzy linguistic terms within the HFLTS framework (naming it, the T2 HFLTS), while using a new construction method for T2 FS based envelopes for the CLEs provided by DMRs. In our proposal, we employ T2 FSs to handle and appropriately model the inter-uncertainty as well as the intra-uncertainty, along with the hesitation faced by the DMRs. In our setting, we utilize interval type-2 trapezoidal fuzzy numbers (IT2 TrFN) to consider the fact that the DMRs may be certain for an interval for a linguistic term. Envelopes are then computed to be used for further computations to provide the final response.

In this paper, we perform computations directly on the measures of uncertainty of T2 FSs, instead of those on indices. This brings out the novelty in our work, which lies in the fact that using IT2 TrFNs for linguistic terms along with computations on uncertainty measures of FSs instead of indices would provide robust and reliable results with good precision. It is expected that our method will further reduce the loss of information while giving good results.

In summary, the contributions of this paper are highlighted below:

1. This is the first work where T2 FSs are introduced in the HFLTS framework to give T2 HFLTS which models inter-uncertainty, intra-uncertainty and hesitation all considered together;
2. A new method to compute fuzzy entropy for T2 HFLTS, based on fuzziness of individual terms is proposed;
3. A novel technique of computing T2 fuzzy envelopes of T2 HFLTS is introduced;
4. A LDM model using T2 HFLTS is provided;
5. The use of IT2 TrFNs is proposed for LDM using T2 HFLTS;
6. A unique priority weightage and expertise based scoring function is proposed.

Rest of the paper is structured as follows: Section II revisits the concepts required to understand our proposed work. In Section III, we introduce some basic concepts and properties of T2 HFLTSs. In Section IV, we discuss about our proposed framework in detail, followed by Section V, wherein we present the working of our model on a supplier performance evaluation (SPE) problem. Section VI presents few comparative studies to demonstrate the novelty of our proposed method. Lastly, we conclude the paper with presenting some discussions and conclusions in Section VII.

## II. PREFATORY KNOWLEDGE

In this section, we revisit some existing information which is required to understand the proposed work.

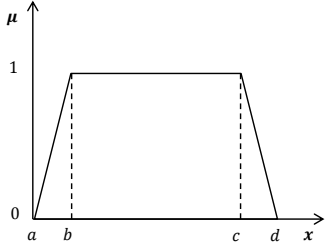

Fig. 1. A T1 FN

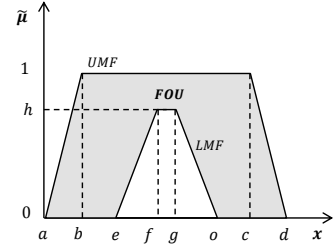

Fig. 2. An IT2 Tr FN

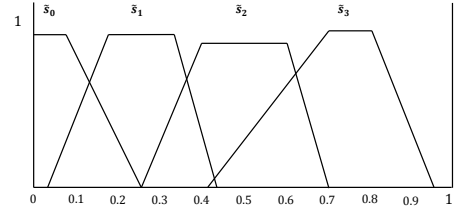

Fig. 3. An HFLTS

## A. Fuzzy Sets: Type-1 and Type-2 [11]

A T1 FS is defined on each element belonging to a set with some degree of membership. Let $X$ be a set of elements, the T1 FS $A$ defined on $X$, assigns graded memberships to every $x \in X$, based on a membership function (MF) denoted by $\mu_A(x) \in [0,1]$. Formally,

$$A = \{(x, \mu_A(x)) | \forall x \in X, \mu_A(\cdot) \in [0,1]\} \quad (1)$$

A fuzzy number (FN) is a special type of FS that satisfies the following properties [18].
1) The FS is normal, i.e. the FS strictly possesses the maximum height of 1.
2) The $\alpha$-cut of the FS must be a closed interval $\forall \alpha \in (0,1]$. The $\alpha$-cut of a FS is defined as the set of all the elements with membership degree $\alpha$ or higher.
3) The support of the FS, i.e., the set of all elements that have non-zero membership is bounded.

A trapezoidal T1 FN is shown in Fig. 1.

On the other hand, T2 FSs assign a secondary degree of membership ($\mu_{\tilde{A}}$) in addition to a primary degree of membership ($J_x$) on the elements of the set in consideration. Consider the same set $X$, for which the T2 FS is denoted by $\tilde{A}$, whose MF is characterized by $\mu_{\tilde{A}}(x, u)$, where $u \in [0,1]$, for every $x \in X$. Formally,

$$\tilde{A} = \{((x, u), \mu_{\tilde{A}}(x, u)) | \forall x \in X, u \in J_x \subseteq [0,1]\}, \quad (2)$$

where $0 \leq \mu_{\tilde{A}}(x, u) \leq 1$.

When, the secondary MF is set to 1 for all elements, the T2 FS becomes an IT2 FS. Therefore,

$$\tilde{A} = \{((x, u), \mu_{\tilde{A}}(x, u) = 1) | \forall x \in X, u \in J_x \subseteq [0,1]\} \quad (3)$$

Fig. 2 demonstrates an IT2 TrFN, $\tilde{A}$. Notice that the shaded region, called the $FOU$ in the figure conveys the uncertainty in the IT2 FS, and uniquely determines the same. The $FOU$ comprises of an upper MF ($UMF$) and a lower MF ($LMF$). All the three are formally defined as follows:

$$FOU(\tilde{A}) = \{(x, u) : x \in X, u \in [\underline{\mu}_{\tilde{A}}(x), \overline{\mu}_{\tilde{A}}(x)]\} \quad (4)$$

$$\underline{\mu}_{\tilde{A}}(x) = \inf\{u | u \in [0,1], \mu_{\tilde{A}}(x, u) > 0\}, \forall x \in X \quad (5)$$

$$\overline{\mu}_{\tilde{A}}(x) = \sup\{u | u \in [0,1], \mu_{\tilde{A}}(x, u) > 0\}, \forall x \in X \quad (6)$$

Eq. (5) denotes the $LMF$ whereas Eq. (6) denotes the $UMF$ for an IT2 TrFN $\tilde{A}$.

IT2 FSs have experienced a recent boost in research within various domains such as similarity measures [19], DM [17:20], fuzzy control [21], [22], fuzzy clustering [23], image segmentation [24] etc. Some recent works related to IT2 FSs also include: general forms of IT2 FSs [25], parameter adaptation [26], IT2 FSs based predictive model [27], etc.

## B. CLEs and HFLTSs

It is now obvious that CLEs convey richer linguistic information which is closer to human cognition. These CLEs, denoted by $ll$, are linguistic expressions or phrases that are generated with the help of a context-free grammar (CGF), $G_H$ whose production rules are defined in an extended Backus-Naur form. But these linguistic expressions must first be transformed into HFLTSs for further processing. Let us first define HFLTSs as given in [7].

**Definition 1 [7]:** Let $S$ be a linguistic term set (LTS) such that $S = \{s_0, \ldots, s_g\}$. Then $H_S = \{s_{\alpha_1}, \ldots s_{\alpha_l}\}$ is an HFLTS which is defined as an ordered finite subset of the consecutive linguistic terms of $S$. $g + 1$ is called the cardinality of $H_S$. An HFLTS is depicted in Fig. 3.

**Definition 2 [7]:** Let $E_{G_H}$ be the transformation function which transforms CLEs $ll$ obtained by $G_H$ into the HFLTS $H_S$ defined on the LTS $S$. Then $E_{G_H}$ is formally defined as follows:

$$E_{G_H}: ll \rightarrow H_S \quad (7)$$

The transformation rules used by $E_{G_H}$ are given below:
1) $E_{G_H}(s_i) = \{s_i / s_i \in S\}$
2) $E_{G_H}(less\ than\ s_i) = \{s_j / s_j \in S \text{ and } s_j \leq s_i\}$
3) $E_{G_H}(more\ than\ s_i) = \{s_j / s_j \in S \text{ and } s_j \geq s_i\}$
4) $E_{G_H}(between\ s_i\ and\ s_j) = \{s_k / s_k \in S \text{ and } s_i \leq s_k \leq s_j\}$.

## C. Entropies

Wei et al. [28] introduced some measures of uncertainty for extended HFLTSs (EHFLTS). Since HFLTSs are special cases of EHFLTSs, the same axiomatic definitions were used to compute the T2 fuzzy envelopes in [17].

The three entropies namely, comprehensive entropy ($E_c$), fuzzy entropy ($E_f$) and hesitant entropy ($E_h$) along with an importance parameter ($\beta$) were used in the computation process, as defined below.

Assume the HFLTS $H_S = \{s_{\alpha_1}, \ldots s_{\alpha_l}\}$ defined on LTS $S = \{s_0, \ldots, s_g\}$, for all the following definitions.

**Definition 3 [17]:** For the HFLTS $H_S$, a fuzzy entropy measure $E_f(H_S)$ is defined as follows:

$$E_f(H_S) = \frac{1}{l}\sum_{i=1}^{l} 4\frac{I(s_{\alpha_i})}{g}\left(1 - \frac{I(s_{\alpha_i})}{g}\right), \quad (8)$$

where, $I(s_{\alpha_k})$ gives the index of the linguistic term $s_{\alpha_k}$ in the LTS $S$.

**Definition 4 [17]:** For the LTS $S$ and HFLTS $H_S$ defined on $S$, a measure of hesitant entropy $E_h(H_S)$ is defined as follows:

$$E_h(H_S) = \frac{1}{g}\eta(H_S), \quad (9)$$

$$\eta(H_S) = \frac{2}{l(l-1)}\sum_{i=1}^{l-1}\sum_{j=i+1}^{l}\left(I(s_{\alpha_j}) - I(s_{\alpha_i})\right) \quad (10)$$

An importance degree denoted as, $\beta$ is used to decide the importance of hesitance of the given HFLTS, depending on the CLEs. It is defined below.

**Definition 5 [17]:** For the three CLEs of the form "less than ...", "more than ..." and "between ... and ...", the definitions of $\beta$ are given below. For $i, j \in [0, g]$:

$$\beta\left(E_{G_H}(\text{more than } s_i)\right) = \frac{1}{2}\cos\frac{\pi}{g}i + \frac{1}{2} \quad (11)$$

$$\beta\left(E_{G_H}(\text{less than } s_i)\right) = \frac{1}{2}\sin\left(\frac{\pi}{g}i - \frac{\pi}{2}\right) + \frac{1}{2} \quad (12)$$

$$\beta\left(E_{G_H}(\text{between } s_i \text{ and } s_j)\right) = \begin{pmatrix}\frac{1}{2}\cos\frac{\pi}{g}i + \\ \frac{1}{2}\sin\left(\frac{\pi}{g}i - \frac{\pi}{2}\right)\end{pmatrix} \quad (13)$$

**Definition 6 [17]:** For the HFLTS $H_S$, the comprehensive entropy $E_c(H_S)$ is defined as follows:

$$E_c(H_S) = \frac{E_f(H_S) + \beta(H_S)E_h(H_S)}{1 + \beta(H_S)E_h(H_S)}, \quad (14)$$

where $\beta \in [0,1]$ varies according to the HFLTS in question. Smaller values of $\beta$ indicate lesser hesitancy within the HFLTS.

In addition to these, Wu and Mendel described the fuzzy entropy measure for IT2 FSs in [29]. It is defined as follows.

**Definition 7 [29]:** A general measure of fuzziness (entropy) of the T1 FS $A$, $f(A)$ is defined as:

$$f(A) = h\left(\sum_{i=1}^{N} g(\mu_A(x_i))\right), \quad (15)$$

where, $h$ is a monotonically increasing function from $R^+$ to $R^+$, and, $g: [0,1] \rightarrow R^+$ is a function associated with each $x_i$. Also, the following properties should be fulfilled by $g$:
P1: $g(0) = g(1) = 0$
P2: $g(0.5)$ is a unique maximum of $g$, and,
P3: $g$ must be monotonically increasing on $[0,0.5]$ and monotonically decreasing on $[0.5,1]$.

**Definition 8 [29]:** The fuzziness of the IT2 FS $\tilde{A}$ is the union of the fuzziness of all its embedded T1 FSs $A_e$, i.e.

$$\tilde{F}_{\tilde{A}_e} \equiv \bigcup_{\forall A_e} f(A_e) = [f_l(\tilde{A}), f_r(\tilde{A})], \quad (16)$$

where,

$$f_l(\tilde{A}) = \min_{\forall A_e} f(A_e), \quad (17)$$

$$f_r(\tilde{A}) = \max_{\forall A_e} f(A_e), \quad (18)$$

and, $f(A_e)$ satisfies Definition 7.

**Theorem 1 [29]:** Let $A_{e1}$ be defined as

$$\mu_{A_{e1}}(x) = \begin{cases} \bar{\mu}_{\tilde{A}}(x), \bar{\mu}_{\tilde{A}}(x) \text{ is further away from 0.5 than } \underline{\mu}_{\tilde{A}}(x) \\ \underline{\mu}_{\tilde{A}}(x), \text{otherwise} \end{cases} \quad (19)$$

and $A_{e2}$ be defined as

$$\mu_{A_{e1}}(x) = \begin{cases} \bar{\mu}_{\tilde{A}}(x), \text{both } \bar{\mu}_{\tilde{A}}(x) \text{ and } \underline{\mu}_{\tilde{A}}(x) \text{ are below 0.5} \\ \underline{\mu}_{\tilde{A}}(x), \text{both } \bar{\mu}_{\tilde{A}}(x) \text{ and } \underline{\mu}_{\tilde{A}}(x) \text{ are above 0.5} \\ 0.5, \text{otherwise} \end{cases} \quad (20)$$

Then, Eq. (17) and Eq. (18) can be computed as

$$f_l(\tilde{A}) = f(A_{e1}) \quad (21)$$

$$f_r(\tilde{A}) = f(A_{e2}), \quad (22)$$

where, $f(A)$ follows Eq. (15).

### D. Type-2 Fuzzy envelope for HFLTSs

Liu et al. proposed the concept of T2 fuzzy envelopes for HFLTSs in [17]. Their idea was to develop envelopes from the HFLTSs which are defined on LTSs with semantics of T1 FSs. We briefly describe the steps followed to obtain the T2 fuzzy envelopes for an HFLTS $H_S = \{s_{i=\alpha_1}, \ldots s_{j=\alpha_l}\}$ defined on LTS $S = \{s_0, \ldots, s_g\}$, such that $s_k \in S = \{s_0, \ldots, s_g\}, k \in \{i, \ldots, j\}$:

1. Compute T1 fuzzy envelope for the $H_S$ following the steps given below:
   i. Let the end points of the linguistic terms to aggregate in $H_S$ be $T = \{a_L^i, a_M^i, a_M^{i+1}, \ldots, a_M^j, a_R^j\}$, following the partitions given in [30]. The semantics of the linguistic terms are defined by T1 TFNs.
   ii. Parameters of aggregated FS are characterized by the end points of a T1 TrFN $F_{H_S} = T(a, b, c, d)$, which are calculated as:

   $a = \min\{a_L^i, a_M^i, a_M^{i+1}, \ldots, a_M^j, a_R^j\} = a_L^i$,
   $d = \max\{a_L^i, a_M^i, a_M^{i+1}, \ldots, a_M^j, a_R^j\} = a_R^j$,
   $b = OWA_{W^s}(a_M^i, a_M^{i+1}, \ldots, a_M^j)$,
   $c = OWA_{W^t}(a_M^i, a_M^{i+1}, \ldots, a_M^j)$,

   where $s, t = 1,2, s \neq t$ or $s = t$.
   iii. Compute the $OWA$ weights as given below:

**Definition 9 [31]:** Let any $\alpha \in [0,1]$, then weights $W^1 = (w_1^1, w_2^1, \ldots, w_n^1)$ are defined as:
$w_1^1 = \alpha, w_2^1 = \alpha(1-\alpha), w_3^1 = \alpha(1-\alpha)^2, \ldots, w_{n-1}^1 = \alpha(1-\alpha)^{n-2}, w_n^1 = \alpha(1-\alpha)^{n-1}$.
Also, weights $W^2 = (w_1^2, w_2^2, \ldots, w_n^2)$ are defined as: $w_1^2 = \alpha^{n-1}, w_2^2 = (1-\alpha)\alpha^{n-2}, w_3^2 = (1-\alpha)\alpha^{n-3}, \ldots, w_{n-1}^2 = (1-\alpha)\alpha, w_n^2 = 1 - \alpha$.

iv. The T1 fuzzy envelope, $F_{H_S}$ is hence computed for the HFLTS $H_S$ as $F_{H_S} = T(a,b,c,d)$, the MF of which is characterized as $F_{H_S}(x), \forall x \in X$.

2. $F_{H_S}$ computed in the previous step is the *UMF* of the T2 fuzzy envelope, $\tilde{F}_{H_S}$. Therefore, for the *FOU* of the envelope, $\overline{\mu}_{\tilde{F}_{H_S}}(x) = F_{H_S}$. For the *LMF*, determine comprehensive entropy, $E_c(H_S)$ for $H_S$ using Eq. (8). Then, $\underline{\mu}_{\tilde{F}_{H_S}}(x) = \max\{0, F_{H_S}(x) - E_c(H_S)\}, \forall x \in X$. $\tilde{F}_{H_S}$ is then uniquely determined by the just computed *FOU*.

## III. TYPE-2 FUZZY SETS BASED HESITANT FUZZY LINGUISTIC TERM SETS

It has been proven time and again that T1 FSs are not capable enough to model linguistic uncertainties of a word. The lesser degrees of freedom within a T1 FS are unable to model the word and expert-level uncertainties together. Hence, T2 FSs are scientifically correct to model linguistic uncertainties. This is because the higher degrees of freedom of T2 FSs are capable of capturing the word and expert-level uncertainties well.

Therefore, to make the HFLTS framework robust enough to handle both the word and expert-level uncertainties along with the hesitation that the experts experience, T2 FSs must be employed for information representation. In this section, we formally define a T2 FS based HFLTS (T2 HFLTS) and give the related concepts for their usage in DM problems.

**Definition 10:** Let $\tilde{S}$ be an LTS such that, $\tilde{S} = \{\tilde{s}_0, \ldots, \tilde{s}_g\}$, where each linguistic term $\tilde{s}_i, i \in [0,g]$ is represented using an IT2 TrFN. Then, T2 HFLTS, $\tilde{H}_{\tilde{S}}$ is defined as an ordered finite subset of the consecutive linguistic terms of $\tilde{S}$. $g+1$ is the cardinality of the LTS $\tilde{S}$.

An LTS, $\tilde{S}$ based on T2 FSs is depicted in Fig 4. One T2 HFLTS defined on $\tilde{S}$ is shown in Fig 5.

Based on the above definition, we can define the types of T2 HFLTS, as follows:
1) Empty T2 HFLTS: $\tilde{H}_{\tilde{S}} = \{\}$,
2) Full T2 HFLTS: $\tilde{H}_{\tilde{S}} = \tilde{S}$,
3) Any other HFLTS: $\tilde{H}_{\tilde{S}} = \{\tilde{s}_i \in \tilde{S}\}$.

It is very well known that an IT2 FS can be viewed as a collection of multiple embedded T1 FSs [32]. This is true in the case of linguistic information because the T2 FS based representation of a linguistic term is often seen as a collection of responses by different experts. Keeping this point in mind, we make the following remark.

**Remark 1:** HFLTSs based on T1 FSs can be viewed as special cases of T2 HFLTSs. Hence, a T2 HFLTS can analogously be represented as a collection of multiple embedded HFLTSs based on T1 FSs, as shown in Fig 6. This fact can be used to extend operations and computations on T2 HFLTSs, by first performing operations and computations on the embedded HFLTSs collectively. However, a T2 HFLTS may contain huge numbers of embedded HFLTSs, which would make it cumbersome to compute with all combinations.

We now define some computations and operations to be performed on T2 HFLTSs.

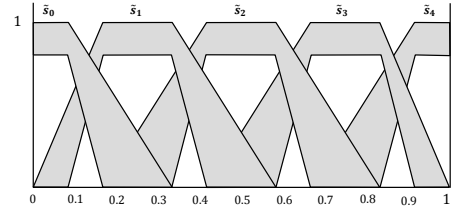
Fig. 4. A T2 FS based LTS

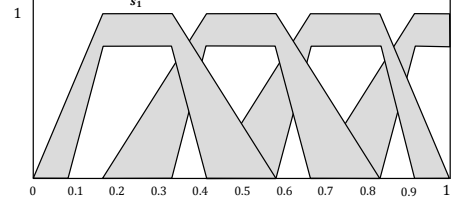
Fig. 5. A T2 HFLTS defined on $\tilde{S}$

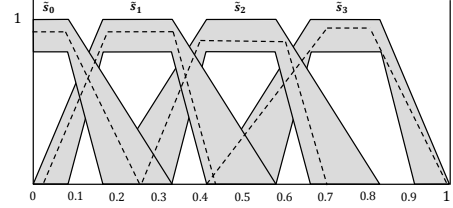
Fig. 6. An embedded HFLTS within a T2 HFLTS

**Definition 11:** The complement or the negative of a T2 HFLTS $\tilde{H}_{\tilde{S}} = \{\tilde{s}_{\alpha_1}, \tilde{s}_{\alpha_2}, \ldots \tilde{s}_{\alpha_l}\}$, defined on LTS $\tilde{S} = \{\tilde{s}_0, \ldots, \tilde{s}_g\}$ is defined as

$$\overline{\tilde{H}_{\tilde{S}}} = \{\tilde{s}_{g-i} | \tilde{s}_i \in \tilde{H}_{\tilde{S}}\}.$$

**Definition 12:** The union of two T2 HFLTSs, $\tilde{H}_{\tilde{S}}^1$ and $\tilde{H}_{\tilde{S}}^2$ is defined as

$$\tilde{H}_{\tilde{S}}^1 \cup \tilde{H}_{\tilde{S}}^2 = \{\tilde{s}_i | \tilde{s}_i \in \tilde{H}_{\tilde{S}}^1 \text{ or } \tilde{s}_i \in \tilde{H}_{\tilde{S}}^2\}.$$

**Definition 13:** The intersection of two T2 HFLTSs, $\tilde{H}_{\tilde{S}}^1$ and $\tilde{H}_{\tilde{S}}^2$ is defined as

$$\tilde{H}_{\tilde{S}}^1 \cap \tilde{H}_{\tilde{S}}^2 = \{\tilde{s}_i | \tilde{s}_i \in \tilde{H}_{\tilde{S}}^1 \text{ and } \tilde{s}_i \in \tilde{H}_{\tilde{S}}^2\}.$$

Some relevant properties of T2 HFLTSs are presented below. However, the proofs of the same being straightforward have been omitted from this article.
Let $\tilde{H}_{\tilde{S}}^1, \tilde{H}_{\tilde{S}}^2$ and $\tilde{H}_{\tilde{S}}^3$ be three T2 HFLTSs.
1) Involution
$$\overline{\overline{\tilde{H}_{\tilde{S}}^1}} = \tilde{H}_{\tilde{S}}^1.$$
2) Commutative
$$\tilde{H}_{\tilde{S}}^1 \cup \tilde{H}_{\tilde{S}}^2 = \tilde{H}_{\tilde{S}}^2 \cup \tilde{H}_{\tilde{S}}^1,$$
$$\tilde{H}_{\tilde{S}}^1 \cap \tilde{H}_{\tilde{S}}^2 = \tilde{H}_{\tilde{S}}^2 \cap \tilde{H}_{\tilde{S}}^1.$$
3) Associative
$$\tilde{H}_{\tilde{S}}^1 \cup (\tilde{H}_{\tilde{S}}^2 \cup \tilde{H}_{\tilde{S}}^3) = (\tilde{H}_{\tilde{S}}^1 \cup \tilde{H}_{\tilde{S}}^2) \cup \tilde{H}_{\tilde{S}}^3$$
$$\tilde{H}_{\tilde{S}}^1 \cap (\tilde{H}_{\tilde{S}}^2 \cap \tilde{H}_{\tilde{S}}^3) = (\tilde{H}_{\tilde{S}}^1 \cap \tilde{H}_{\tilde{S}}^2) \cap \tilde{H}_{\tilde{S}}^3.$$
4) Distributive
$$\tilde{H}_{\tilde{S}}^1 \cap (\tilde{H}_{\tilde{S}}^2 \cup \tilde{H}_{\tilde{S}}^3) = (\tilde{H}_{\tilde{S}}^1 \cap \tilde{H}_{\tilde{S}}^2) \cup (\tilde{H}_{\tilde{S}}^1 \cap \tilde{H}_{\tilde{S}}^3),$$
$$\tilde{H}_{\tilde{S}}^1 \cup (\tilde{H}_{\tilde{S}}^2 \cap \tilde{H}_{\tilde{S}}^3) = (\tilde{H}_{\tilde{S}}^1 \cup \tilde{H}_{\tilde{S}}^2) \cap (\tilde{H}_{\tilde{S}}^1 \cup \tilde{H}_{\tilde{S}}^3).$$

Based on these basic definitions, we now define the process of LDM using T2 HFLTS in the next section.

IV. LINGUISTIC DECISION MAKING USING T2 HFLTS

The existing frameworks of HFLTS cannot be used to model the word and expert-level uncertainties namely, the intra-uncertainty and inter-uncertainty respectively. It must also be pointed out that in [17], the computations made on indices for the fuzzy entropy do not reflect the true 'fuzziness' inherent in the linguistic terms in the considered HFLTS. Therefore, to overcome these problems and to introduce other concepts, we propose the framework and the concepts related to T2 HFLTS in this Section.

A T2 FS based linguistic term set $\tilde{S}$ with the terms $\tilde{S} = \{very\ poor, poor, moderate, good, very\ good\}$ or $\tilde{S} = \{\tilde{s}_0: VP, \tilde{s}_1: P, \tilde{s}_2: M, \tilde{s}_3: G, \tilde{s}_4: VG\}$ is shown in Fig. 4.

We first present the related concepts required for computing with $\tilde{H}_{\tilde{S}}$. In Section IV-A through Section IV-E, we present a linguistic multi-criteria group decision making (MCGDM) framework based on T2 HFLTSs. This framework includes a group of DMRs who assess the alternatives given to them, based on few criteria. The DMRs are provided with an LTS, fixed a priori through experts' knowledge. The DMRs then respond with phrases or CLEs that closely resemble their actual assessment in natural language, with terms drawn from the LTS.

These assessments, whether CLEs or single terms are represented by IT2 FSs drawn from the LTS. Then the T2 fuzzy envelopes are computed for all the CLEs in the assessments. Based on this representation of CLEs, the T2 fuzzy envelopes are aggregated for every alternative, corresponding to every criterion, for every DMR. Then the resultant IT2 FSs are ranked on the basis of their centroids. Once the ranking for alternatives for every DMR is known, the unique ranking criteria presented in Section IV-E are used to obtain the final ranking of the alternatives.

The LDM framework mentioned above can be broadly divided into five steps. A pictorial representation of the same is given in Fig. 7. All the steps are briefly described below.

1) *Survey conduction and response generation:* A survey is conducted by providing an LTS with IT2 TrFNs as semantics, to the DMRs. This LTS is fixed a priori with the help of experts' knowledge. The DMRs respond with their choice of phrases (CLEs) or terms, drawn from the given LTS.

2) *Transformation of CLE to T2 HFLTS:* For any CLE, it is transformed into the corresponding T2 HFLTS using the transformation function $E_{G_H}$, as given in Definition 2. Note that the responses with only single terms drawn from the given LTS do not require any transformation.

3) *Envelope computation:* The linguistic terms in the obtained T2 HFLTS from the previous step are clubbed together to form the T2 fuzzy envelope. However, the fuzzy entropy for the T2 HFLTS is computed based on the Definition 8. The detailed procedure for the same is provided in Section IV-C.

4) *Aggregation:* The responses of each DMR, whether a single term or a T2 fuzzy envelope, are then aggregated corresponding to every criteria for every alternative, to give one T2 FS.

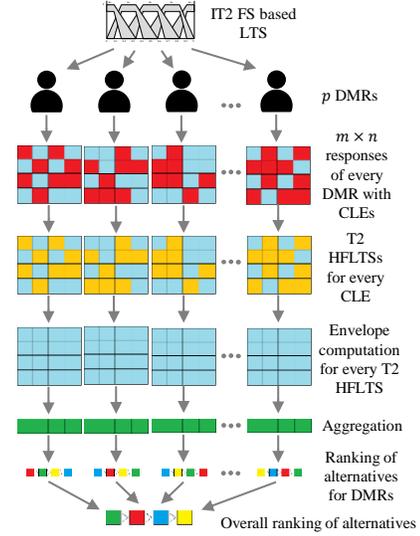

Fig. 7. LDM process with multiple DMRs using T2 HFLTS

5) *Ranking:* These T2 FSs are then ranked based on a new ranking formula discussed later in Section IV-E.

We now discuss each of the above steps in detail in the following subsections.

A. *Details of survey conduction and response generation*

With the help of experts' knowledge, an LTS is fixed a priori, to be used by the DMRs for their responses. An LTS, $\tilde{S} = \{\tilde{s}_0, \ldots, \tilde{s}_g\}$ (*see* Fig 4.) with IT2 TrFNs as semantics for linguistic terms is decided.

Suppose there are $p$ DMRs denoted as $\mathfrak{D} = \{D_1, D_2, \ldots, D_p\}$, who must respond with CLEs or terms for the performance of $m$ alternatives denoted as $\mathfrak{A} = \{A_1, A_2, \ldots, A_m\}$, based on $n$ criteria denoted as $\mathfrak{C} = \{C_1, C_2, \ldots, C_n\}$. This results in $p \times m \times n$ responses to deal with and generate the final solution.

The responses that are phrases or CLEs, $ll$ including multiple terms drawn from $\tilde{S}$, are generated with the help of the production rules defined within the CFG, $G_H$. There are three types of CLEs that we consider the CFG $G_H$ to generate. These are follows:

1. $ll =$ between $\tilde{s}_i$ and $\tilde{s}_j, \tilde{s}_i, \tilde{s}_j \in \tilde{S}$
2. $ll =$ less than $\tilde{s}_i, \tilde{s}_i \in \tilde{S}$
3. $ll =$ more than $\tilde{s}_i, \tilde{s}_i \in \tilde{S}$

It is crucial to mention that the responses that include single terms are not treated as HFLTSs, but rather as individual terms only. This greatly reduces the computations in cases where the DMRs are not very hesitant about their responses. It is crucial to point here that this also helps during the final aggregation phase as the individual linguistic terms are already IT2 TrFNs, and hence do not require unnecessary computations. For e.g., in [17], an individual term say, '$s_i \in S$' is considered as the CLE 'between $s_i$ and $s_i$', which leads to an unnecessary computation of the envelope for the CLE. Our consideration of T2 HFLTSs eliminates this.

B. *The process of transformation of CLE to T2 HFLTS*

This step transforms any of the obtained CLEs to a T2 HFLTS say, $\tilde{H}_{\tilde{S}}$ following Definition 2, using the transformation function, $E_{G_H}$. Sample T2 HFLTS can be viewed as any of the three given in Fig. 8.

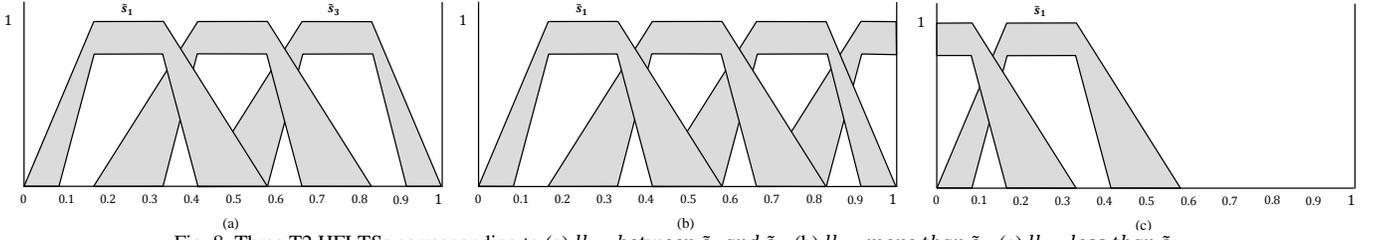

Fig. 8. Three T2 HFLTSs corresponding to (a) $ll = between\ \tilde{s}_1\ and\ \tilde{s}_3$, (b) $ll = more\ than\ \tilde{s}_1$, (c) $ll = less\ than\ \tilde{s}_1$.

These T2 HFLTSs are obtained by applying transformation rules given in Definition 2 on the CLEs as shown below:

(a) $E_{G_H}(ll = between\ \tilde{s}_1\ and\ \tilde{s}_3) = \{\tilde{s}_1, \tilde{s}_2, \tilde{s}_3\}$,
(b) $E_{G_H}(ll = more\ than\ \tilde{s}_1) = \{\tilde{s}_1, \tilde{s}_2, \tilde{s}_3, \tilde{s}_4\}$,
(c) $E_{G_H}(ll = less\ than\ \tilde{s}_1) = \{\tilde{s}_0, \tilde{s}_1\}$.

Fig. 8(a)-(c) respectively correspond to the T2 HFLTSs of the CLEs, which are obtained above.

### C. Computation of the T2 fuzzy envelopes from T2 HFLTS

As already discussed in Section I, T2 fuzzy envelopes provide better representations than any of the previous counterparts. However, the computations for the same defined in [17] do not consider T2 HFLTSs in their framework. Also, the computations of fuzzy entropies for any HFLTS were made on the indices of the linguistic terms. Such computations do not reflect the true amount of fuzziness present in an HFLTS. Therefore, we propose a new method to compute the T2 fuzzy envelope for a T2 HFLTS, $\widetilde{H}_{\tilde{S}} = \{\tilde{s}_{i=\alpha_1}, \tilde{s}_{i+1=\alpha_2}, \ldots \tilde{s}_{j=\alpha_l}\}$. First, T1 fuzzy envelopes are computed separately for both the $UMF$ and the $LMF$ of the terms in the T2 HFLTS. The $UMF$ hence obtained acts as the $UMF$ of the final T2 fuzzy envelope. Then the entropies are obtained for the linguistic terms and hence, the T2 HFLTSs. These are then used along with the T1 fuzzy envelopes computed to calculate that LMF of the T2 fuzzy envelope.

The detailed steps to compute T2 fuzzy envelopes are given below:

1. Recall that every linguistic term $\tilde{s}_k \in \widetilde{H}_{\tilde{S}} \subseteq \tilde{S}$, $i \in [i,j]$, is represented by an IT2 TrFN, which in turn is uniquely determined by its $FOU$. Let every term $\tilde{s}_i$ be represented as $T\{(a,b,c,d),(e,f,g,o),h\}$ as shown in Fig. 2. Therefore, $UMF = T_U(a,b,c,d)$ and $LMF = T_L(e,f,g,o)$, with height $h$.

2. The elements to aggregate are:
   i. For $UMF$: $T_{UMF} = \{a^i, b^i, c^i, a^{i+1}, d^i, \ldots, d^{j-1}, c^j, d^j\}$
   ii. For $LMF$: $T_{LMF} = \{e^i, f^i, g^i, e^{i+1}, o^i, \ldots, o^{j-1}, g^j, o^j\}$

Let the MFs of the T1 fuzzy envelopes, $F_{\widetilde{H}_{\tilde{S}}}^{UMF}$ and $F_{\widetilde{H}_{\tilde{S}}}^{LMF}$, be denoted by $T_U(a_U, b_U, c_U, d_U)$ and $T_L(e_L, f_L, g_L, o_L)$ respectively. To calculate the T1 fuzzy envelopes compute the following:

$a_U = \min\{a^i, b^i, c^i, a^{i+1}, d^i, \ldots, d^{j-1}, c^j, d^j\} = a^i$,
$d_U = \max\{a^i, b^i, c^i, a^{i+1}, d^i, \ldots, d^{j-1}, c^j, d^j\} = d^j$,
$b_U = OWA_{W^s}\left(\frac{(b^i+c^i)}{2}, \frac{(b^{i+1}+c^{i+1})}{2}, \ldots, \frac{(b^j+c^j)}{2}\right)$,
$c_U = OWA_{W^t}\left(\frac{(b^i+c^i)}{2}, \frac{(b^{i+1}+c^{i+1})}{2}, \ldots, \frac{(b^j+c^j)}{2}\right)$,

where $s, t = 1,2$, $s \neq t$ or $s = t$.

Similarly:
$e_L = \min\{e^i, f^i, g^i, e^{i+1}, o^i, \ldots, o^{j-1}, g^j, o^j\} = e^i$,
$f_L = \max\{e^i, f^i, g^i, e^{i+1}, o^i, \ldots, o^{j-1}, g^j, o^j\} = o^j$,
$g_L = OWA_{W^s}\left(\frac{(f^i+g^i)}{2}, \frac{(f^{i+1}+g^{i+1})}{2}, \ldots, \frac{(f^j+g^j)}{2}\right)$,
$o_L = OWA_{W^t}\left(\frac{(f^i+g^i)}{2}, \frac{(f^{i+1}+g^{i+1})}{2}, \ldots, \frac{(f^j+g^j)}{2}\right)$,

where $s, t = 1,2$, $s \neq t$ or $s = t$.

The $OWA$ operator and the weights are calculated as given in [30] and reviewed in Section II-D. Note that the points $b$, $c$, $f$ and $g$ are given equal weightage during the computations as they have equal membership functions, i.e. $\overline{\mu}(b) = \overline{\mu}(c)$ and $\underline{\mu}(f) = \underline{\mu}(g)$. Now, obtain $F_{\widetilde{H}_{\tilde{S}}}^{UMF}$ and $F_{\widetilde{H}_{\tilde{S}}}^{LMF}$ accordingly.

3. To compute the uncertainty contained within the T2 HFLTS, $\widetilde{H}_{\tilde{S}}$, its entropies are computed. The entropies in question are the fuzzy entropy $E_f(\widetilde{H}_{\tilde{S}})$ and hesitant entropy $E_h(\widetilde{H}_{\tilde{S}})$ which together are used to compute the comprehensive entropy $(E_c(\widetilde{H}_{\tilde{S}}))$. The comprehensive entropy measure is a class of entropy measures that follow certain definitions as given in [28]. A formulation of $E_c(H_S)$ and $E_h(H_S)$ is already discussed in Definition 4 and Definition 6 respectively.

We now introduce a new definition for fuzzy entropy of T2 HFLTSs based on the fuzzy entropy of IT2 FSs, given Definition 8. We first give an axiomatic definition of the proposed fuzzy entropy measure for T2 HFLTSs.

**Definition 14:** For the T2 HFLTS $\widetilde{H}_{\tilde{S}} = \{\tilde{s}_{\alpha_1}, \tilde{s}_{\alpha_2}, \ldots \tilde{s}_{\alpha_l}\}$ defined on the LTS $\tilde{S}$, let $\mathbb{H}(\tilde{S})$ be the set of all the T2 HFLTSs defined on $\tilde{S}$, such that $\widetilde{H}_{\tilde{S}} \in \mathbb{H}(\tilde{S})$. Let $\widetilde{E}_f: \mathbb{H}(\tilde{S}) \to R^+$ be the fuzzy entropy measure that satisfies the following axiomatic requirements. We then say that $\widetilde{E}_f$ is the fuzzy entropy for T2 HFLTSs.

(FE1) $\widetilde{E}_f(\widetilde{H}_{\tilde{S}}) = 0$, if and only if $\widetilde{H}_{\tilde{S}} = \{\tilde{s}_0\}$ or $\widetilde{H}_{\tilde{S}} = \{\tilde{s}_g\}$;

(FE2) $\widetilde{E}_f\left(\{\tilde{s}_{\frac{g}{2}}\}\right)$ is a unique maximum of $\widetilde{E}_f$;

(FE3) Let $\widetilde{H}_{\tilde{S}}^1 = \{\tilde{s}_{\alpha_1}, \tilde{s}_{\alpha_2}, \ldots \tilde{s}_{\alpha_l}\}$ be a T2 HFLTS and $\widetilde{H}_{\tilde{S}}^2$ be another T2 HFLTS obtained by changing the element $\tilde{s}_{\alpha_i}(i = 1,2, \ldots, l)$ in $\widetilde{H}_{\tilde{S}}^1$ to $\tilde{s}_{\alpha_i'}$. If $\left|I(\tilde{s}_{\alpha_i}) - \frac{g}{2}\right| \geq \left|I\left(\tilde{s}_{\alpha_i'}\right) - \frac{g}{2}\right|$, then $\widetilde{E}_f(\widetilde{H}_{\tilde{S}}^1) \leq \widetilde{E}_f(\widetilde{H}_{\tilde{S}}^2)$;

(FE4) $\widetilde{E}_f(\widetilde{H}_{\tilde{S}}) = \widetilde{E}_f(\widetilde{H}_{\tilde{S}})$.

Following the above axiomatic definition of the fuzzy entropy measure for a T2 HFLTS, we propose the definition of a new fuzzy entropy measure for individual T2 HFLTSs. It is given in Definition 15.

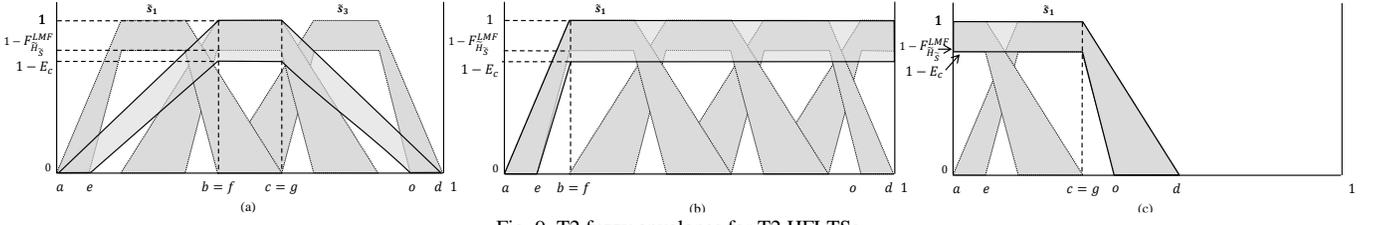

Fig. 9. T2 fuzzy envelopes for T2 HFLTSs.

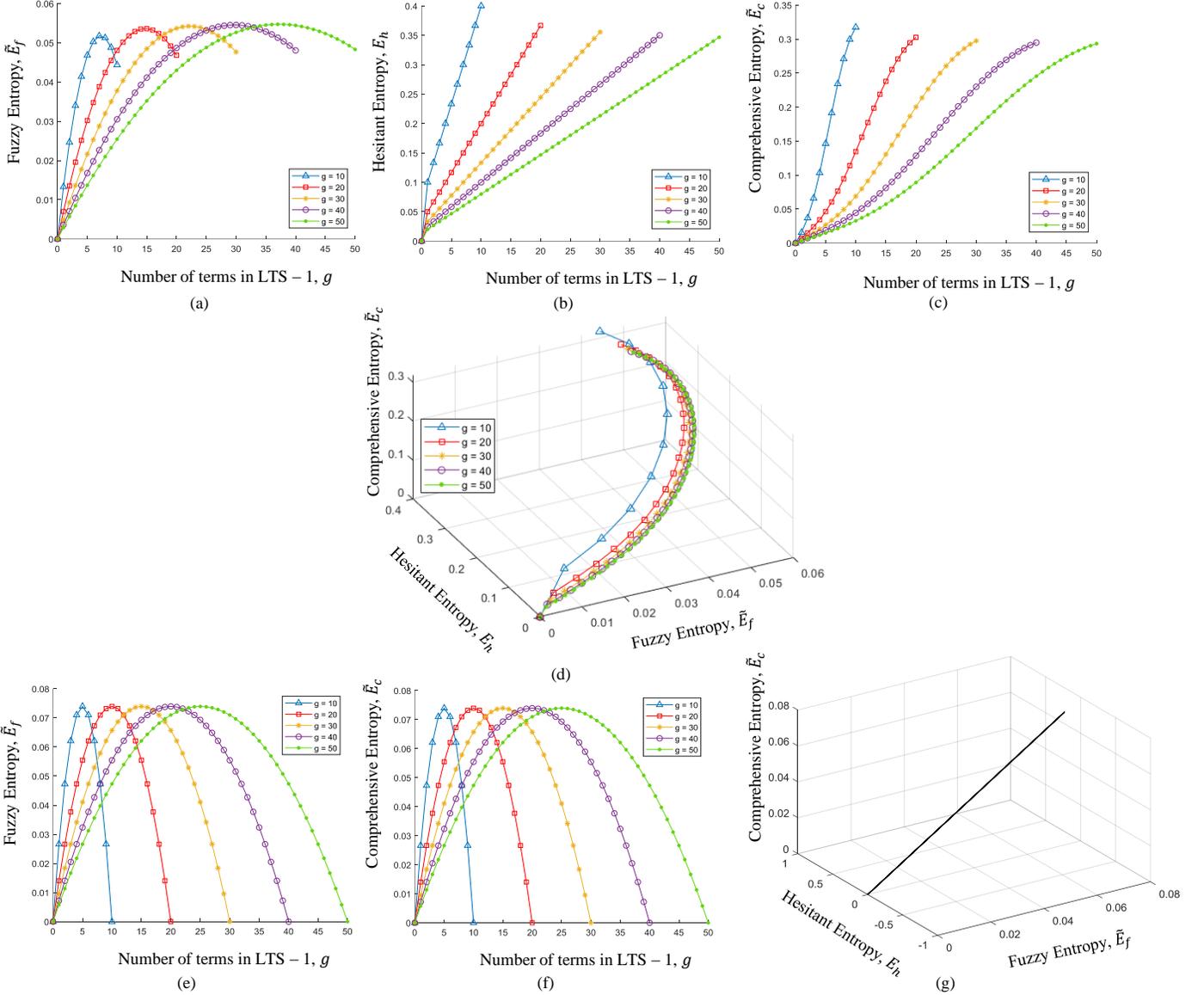

Fig. 10. Results of sensitivity analysis of various entropies for different values of g.

**Definition 15:** Let $\tilde{E}_f$ be defined for the T2 HFLTS $\widetilde{H}_{\tilde{S}}$ as:

$$\tilde{E}_f(\widetilde{H}_{\tilde{S}}) = \frac{1}{l}\sum_{i=1}^{l}\left(4\left(\tilde{F}_{\tilde{s}_{\alpha_i}}\right) \times \frac{I(\tilde{s}_{\alpha_i})}{g}\left(1 - \frac{I(\tilde{s}_{\alpha_i})}{g}\right)\right) \quad (23)$$

where, $\tilde{F}_{\tilde{s}_{\alpha_i}}$ is the IT2 fuzzy entropy of the T2 FS based linguistic term $\tilde{F}_{\tilde{s}_{\alpha_i}}$.

Our definition of $\tilde{E}_f$, satisfies the properties (FE1)-(FE4). The proof is given in the supplementary materials in Section SM-I. Hence, the comprehensive entropy for T2 HFLTS $\widetilde{H}_{\tilde{S}}$, $\tilde{E}_c(\widetilde{H}_{\tilde{S}})$ is defined as follows

$$\tilde{E}_c(\widetilde{H}_{\tilde{S}}) = \frac{\tilde{E}_f(\widetilde{H}_{\tilde{S}}) + \beta(\widetilde{H}_{\tilde{S}})E_h(\widetilde{H}_{\tilde{S}})}{1 + \beta(\widetilde{H}_{\tilde{S}})E_h(\widetilde{H}_{\tilde{S}})}, \quad (24)$$

where $E_h(\widetilde{H}_{\tilde{S}})$ is computed using Eq. (10) and Eq. (11).

The values of $\beta(\widetilde{H}_{\tilde{S}}) \in [0,1]$ [17] in Eq. (24) are defined differently for different T2 HFLTSs. They are given in Definition 5.

**Note:** It is to be noted here that we adopt the Yager's definition [33] of fuzziness for computing the measure given in Eq. (15). This is because, out of the other existing definitions of the measure $f(A)$, the Yager's definition is a normalized version which converges as the number of points on the universe increases [29]. Also, out of the many other fuzziness measures for IT2 FSs, the method by Wu and Mendel is employed here because it is based on the Mendel-John representation theorem. It is good to note that, the other measures of IT2 fuzziness can be shown as special cases of Wu and Mendel's fuzziness measure.

4. The T2 fuzzy envelope, $\tilde{F}_{\widetilde{H}_{\tilde{S}}}$ for the T2 HFLTS, $\widetilde{H}_{\tilde{S}}$ is computed with the help of the $F_{\widetilde{H}_{\tilde{S}}}^{UMF}$, $F_{\widetilde{H}_{\tilde{S}}}^{LMF}$ and $\tilde{E}_c(\widetilde{H}_{\tilde{S}})$. As already known that the T2 fuzzy envelope is uniquely determined by its $FOU$, the $UMF$ of the $FOU$ for $\tilde{F}_{H_S}$ is the same as $F_{\widetilde{H}_{\tilde{S}}}^{UMF}$. Therefore,

$$\bar{\mu}_{\tilde{F}_{\widetilde{H}_{\tilde{S}}}}(x) = F_{\widetilde{H}_{\tilde{S}}}^{UMF} \quad (25)$$

The $LMF$ of the $FOU$ is computed with the help of $F_{\widetilde{H}_{\tilde{S}}}^{LMF}$ and $\tilde{E}_c$ as given below. $\forall x \in X$,

$$\underline{\mu}_{\tilde{F}_{\widetilde{H}_{\tilde{S}}}}(x) = \max\left\{0, F_{\widetilde{H}_{\tilde{S}}}^{UMF} - \max\left\{\begin{pmatrix}(F_{\widetilde{H}_{\tilde{S}}}^{UMF}(x) - F_{\widetilde{H}_{\tilde{S}}}^{LMF}(x)),\\ \tilde{E}_c(H_S)\end{pmatrix}\right\}\right\} \quad (26)$$

**Remark 2:** The above way of calculating $\underline{\mu}_{\tilde{F}_{\widetilde{H}_{\tilde{S}}}}$ makes sure that the resultant $FOU$ of the T2 fuzzy envelope captures the maximum uncertainty gathered in the process of calculation, while keeping it compact. Since, entropy of a fuzzy set is a measure of fuzziness or vagueness contained in it, the higher the value of entropy the higher is the uncertainty within the fuzzy set. Hence, if the obtained value of $\tilde{E}_c$ is greater than the existing height of the $F_{\widetilde{H}_{\tilde{S}}}^{LMF}$, the uncertainty of the terms within the concerned T2 HFLTS considered together is more than that of the term(s) with higher uncertainty within the same T2 HFLTS considered individually.

Three T2 fuzzy envelopes computed for the T2 HFLTSs in Fig 8 are shown in Fig 9.

Performing a sensitivity analysis on the values of fuzzy, hesitant and comprehensive entropies of T2 HFLTSs for different values of $g$, we obtain interesting results. These results are depicted in the form of line plots Fig 10, and are discussed in details below.

Two cases are considered to perform the sensitivity analysis for the T2 HFLTSs.

*Case 1:* Beginning with an empty T2 HFLTS, it is updated by adding terms that are picked one-by-one incrementally from the LTS.

*Case 2:* Beginning with a T2 HFLTS containing the first term from the LTS and then replacing that term with terms picked incrementally one-by-one from the LTS. The T2 HFLTS in this Case always contains a single element.

**Discussion**

It can be seen from Fig. 9(a) that when Case 1 is considered, the fuzzy entropy, $\tilde{E}_f$ increases up to a certain extent and then decreases when nearing the value of $g$. However, the value of hesitant entropy, $E_h$ is a monotonically increasing function when the number of terms in a T2 HFLTS increases. The comprehensive entropy, $\tilde{E}_c$ on the other hand is a monotonically increasing function in $[0, \frac{g}{2}]$, whereas, it is a monotonically decreasing function in $[\frac{g}{2}, g]$, as can be seen in Fig. 9(c). Fig. 9(d) shows that the comprehensive entropy is a strictly monotonically increasing function with respect to the fuzzy and hesitant entropies for T2 HFLTSs.

When Case 2 is considered, it can be seen from Fig. 9(e) and Fig. 9(f) that both $\tilde{E}_f$ and $\tilde{E}_c$ have identical curves. This is because, when the T2 HFLTS consists of a single term, the importance factor $\beta = 0$. This can be verified from the Definition 5. Again, $\tilde{E}_c$ is a strictly monotonically increasing function with respect to $\tilde{E}_f$ and $E_h$, as shown in Fig. 9(g). It can also be deduced that the curves of $\tilde{E}_c$ for different values of $g$, with respect to $\tilde{E}_f$ and $E_h$ in Case 2, lie superimposed on each other. Therefore, a single curve represents the same for all values of $g$.

*D. Aggregating the responses represented by T2 fuzzy envelopes*

Post to obtaining the T2 fuzzy envelopes, the responses for every alternative corresponding to the criteria are aggregated for every DMR.

Since every response to be aggregated is an IT2 FS, an IT2 FS based aggregation method is chosen. In our work, we make use of the linguistic weighted average (LWA) [34]. It is an extension of the still most widely used form of aggregation, the weighted average. The LWA is generally defined as follows:

$$\tilde{Y}_{LWA} = \frac{\sum_{i=1}^{q} \tilde{X}_i \widetilde{W}_i}{\sum_{i=1}^{q} \widetilde{W}_i} \quad (27)$$

where, $\tilde{X}$ is a set of $q$ IT2 FSs and $\widetilde{W}$ are the corresponding weights, also represented using T2 FSs.

The result of the aggregation process for our proposed decision making framework is given as follows:

$$Agg_i = \tilde{Y}_{LWA}\left(\tilde{F}_{\widetilde{H}_{\tilde{S}_{j,k}}}, cw_k\right), \forall D_i \in \mathfrak{D} \quad (28)$$

Here, $CW = \{cw_1, cw_2, \ldots, cw_m\}$ are the set of criteria weights. Also, Eq. (28) is defined for $\forall j \in \{1,2,\ldots,n\}, \forall k \in \{1,2,\ldots,m\}$.

*E. Details of the ranking process*

$Agg_i$ gives the set of aggregated IT2 FSs corresponding to every alternative in the system for the $i^{th}$ DMR. Now, the final rankings must be generated to find out the optimal alternative as per the assessments provided by the DMRs.

Every DMR is assumed to have a different level of expertise, based on the level of experience of the person. Such expertise of a person must be taken into account considering the fact that assessments coming from better experience provide more reliable results. Keeping this is mind, we propose a unique scoring function for our DM framework below, which considers the assessments provided by each DMR, along with the expertise that every DMR has. This expertise is denoted using a weight vector defined as $\mathfrak{D}_W = \{D_{w_1}, D_{w_2}, \dots, D_{w_p}\}$. As for every weight vector, the entries in $\mathfrak{D}_W$ add up to 1.

The ranking process is divided into two steps as given below.

**Step 1:** *Rank the alternatives individually for every DMR*: Firstly, the aggregated IT2 FSs are ranked to generate the ranking of alternatives corresponding to every DMR. This is achieved by ranking the IT2 FSs, based on their centroids. For this purpose, the centroids are computed using the widely used enhanced Karnik-Mendel (EKM) algorithms [35]. Since the centroid values of IT2 FSs are crisp numbers, it is easy to rank the IT2 FSs associated with them. A brief introduction to the enhanced KM algorithm is given in the supplementary materials in Section SM-II.

**Step 2:** *Generate overall ranking for each alternative*: Let $R$ be a $p \times m$ rank matrix, where every row denotes the ranked alternatives according to centroids, for every DMR. The entries of rank matrix $R$ are denoted as $R_{ij}$, where $i \in \{1,2,\dots,p\}$ and $j \in \{1,2,\dots,m\}$.

A unique scoring function is defined below. This function assigns a score to every alternative which is then used to rank the alternatives. Let the following:

i. $f: \mathfrak{A} \times [1,m] \to \mathbb{N} \cup \{\phi\}$ be a function, such that $f(A_k, r), k \in [1,m]$ returns the frequency of occurrence of the $k^{th}$ alternative, in the $r^{th}$ column of $R$. $f$ returns an empty set if the given alternative does not occur once in the given column of $R$.
ii. $\mathcal{O}: \mathfrak{A} \to \{[1,p] \times [1,m]\}$, be a function such that $\mathcal{O}(A_k)$ returns a set of ordered pairs $(x,y)$, for a given alternative $A_k \in \mathfrak{A}$, such that $x = i$, and $y = j$ when $R_{i,j} = A_k$.
iii. $Y$ denotes the set of all $y$ elements of the ordered pairs $(x,y) \in \mathcal{O}(A_i)$.

Based on the above functions, we now define our proposed scoring function, $\mathfrak{S}: \mathfrak{A} \to \mathbb{R}^+$, is defined by Eq. (29).

$$\mathfrak{S}(A_i) = \frac{\sum_{j \in Y}((\wp_j)C(i,j))}{\sum_{j \in Y}(C(i,j))}, \quad (29)$$

where $\wp_j$ gives the priority weightage function of the rank $j$ such that $\wp: [1,m] \to [1,m]$. Note that $\wp$ is a monotonically decreasing function, for increasing value of $j$. For our proposed framework, $\wp_j = ((m+1) - j)$. Also, the second term in Eq. (29) is defined below:

$$C(i,j) = \frac{1}{f(A_i,j)} \sum_{(k,j) \in \mathcal{O}(A_i)} D_{w_k}. \quad (30)$$

It can be seen clearly that the proposed score function $\mathfrak{S}$, belongs to the family of weighted average. However, the 'weight' in this function, is in itself a function of the weights of the DMRs, the rank of the alternative, and the frequency of the alternative in that rank. The usage of priority weightage ensures that the importance of every rank does not get hidden within the weights of the DMRs.

To obtain the overall ranking of the alternatives, they are ranked based on the descending order of the corresponding values of $\mathfrak{S}$.

**Remark 3:** The LDM model presented above solves problems in the MCGDM environment. It can be well pointed out that, the proposed model can also be utilized for single expert decision making scenarios, i.e., multi-criteria decision making (MCDM) problems. For such a case, the centroid rankings of alternatives obtained in Step 1 of the ranking phase provide the optimal alternative for the given problem.

## V. AN ILLUSTRATIVE EXAMPLE

In this section, we explain the working of the LDM framework using T2 HFLTS proposed in Section IV. The process is demonstrated using an SPE problem defined in Example 1

**Example 1:** A manufacturing company has to set up a new plant in a new city and hence plans to hire a supplier for the new plant. They have prepared a list of five candidate suppliers (alternatives), whom the company needs to assess to find out the best supplier. The company sets up a selection committee consisting of four experts (DMRs), with different levels of expertise to get a proper mix of assessments. These DMRs decide upon four different criteria, $C_1$: long-term relation potential, $C_2$: e-commerce capability, $C_3$: price, quality and delivery and $C_4$: financial stability, along with their respective levels of importance, on which the candidate suppliers are assessed. They also decide upon a LTS, whose semantics being IT2 TrFNs, model the intra-uncertainty as well as the inter-uncertainty appropriately.

The symbolic representations are denoted below:
1. Candidate suppliers: $\mathfrak{A} = \{A_1, A_2, A_3, A_4, A_5\}$
2. Experts: $\mathfrak{D} = \{D_1, D_2, D_3, D_4\}$
3. Expertise levels: $\mathfrak{D}_W = \{0.250, 0.400, 0.150, 0.200\}$
4. Criteria: $\mathfrak{C} = \{C_1, C_2, C_3, C_4\}$
5. LTS, $\tilde{S} = \{\tilde{s}_0: VP, \tilde{s}_1: P, \tilde{s}_2: M, \tilde{s}_3: G, \tilde{s}_4: VG\}$

We now follow the steps discussed in Section IV and solve the above linguistic MCGDM problem.

### 1) Survey conduction and response generation

Each DMR is provided with LTS $\tilde{S}$ which he/she uses to generate their responses. The LTS given in Fig. 4 is chosen.

The above scenario provides $4 \times 5 \times 4$ assessments to be used to find out the optimal supplier for the company. These assessments are given in Table I. Note that the responses include single terms as well as CLEs which closely resemble the inherent thought process of the DMRs, which is close to the natural language.

### 2) Transformation of CLEs to HFLTSs

For all the responses which are CLEs, the transformation function $E_{G_H}$, is used to transform them into T2 HFLTSs. The resultant T2 HFLTSs are given in Table II. The T2 HFLTSs obtained from the responses of DMR $D_1$ are shown in Fig.11(a).

TABLE I RESPONSES BY DMRs IN EXAMPLE 1

| Decision Makers | Criteria; Alternatives | $A_1$ | $A_2$ | $A_3$ | $A_4$ | $A_5$ |
|---|---|---|---|---|---|---|
| $D_1$ | $C_1$ | between $M$ and $VG$ | between $VP$ and $G$ | $G$ | more than $M$ | less than $P$ |
| | $C_2$ | $P$ | $M$ | more than $G$ | between $P$ and $G$ | $P$ |
| | $C_3$ | less than $M$ | more than $G$ | $VG$ | $VP$ | between $VP$ and $P$ |
| | $C_4$ | between $P$ and $VG$ | more than $G$ | less than $M$ | $M$ | $G$ |
| $D_2$ | $C_1$ | $G$ | between $VP$ and $M$ | more than $G$ | less than $P$ | $G$ |
| | $C_2$ | between $G$ and $VG$ | more than $P$ | less than $M$ | $P$ | $G$ |
| | $C_3$ | $M$ | between $P$ and $M$ | $M$ | less than $P$ | more than $G$ |
| | $C_4$ | $P$ | $M$ | less than $P$ | more than $G$ | $VG$ |
| $D_3$ | $C_1$ | less than $P$ | $M$ | more than $G$ | less than $M$ | $P$ |
| | $C_2$ | more than $M$ | more than $G$ | $G$ | $M$ | more than $G$ |
| | $C_3$ | $P$ | between $M$ and $VG$ | more than $P$ | $VP$ | $G$ |
| | $C_4$ | $G$ | less than $P$ | more than $G$ | less than $P$ | $VP$ |
| $D_4$ | $C_1$ | $VG$ | more than $G$ | less than $P$ | $M$ | $G$ |
| | $C_2$ | between $P$ and $M$ | $M$ | $VP$ | more than $G$ | $VG$ |
| | $C_3$ | less than $P$ | more than $G$ | $M$ | $M$ | less than $P$ |
| | $C_4$ | $M$ | between $M$ and $G$ | between $VP$ and $M$ | $P$ | $M$ |

TABLE II T2 HFLTSs OF RESPONSES BY DMRs

| Decision Makers | Criteria; Alternatives | $A_1$ | $A_2$ | $A_3$ | $A_4$ | $A_5$ |
|---|---|---|---|---|---|---|
| $D_1$ | $C_1$ | $\{M,G,VG\}$ | $\{VP,P,M,G\}$ | $G$ | $\{M,G,VG\}$ | $\{VP,P\}$ |
| | $C_2$ | $P$ | $M$ | $\{G,VG\}$ | $\{P,M,G\}$ | $P$ |
| | $C_3$ | $\{VP,P,M\}$ | $\{G,VG\}$ | $VG$ | $VP$ | $\{VP,P\}$ |
| | $C_4$ | $\{P,M,G,VG\}$ | $\{G,VG\}$ | $\{VP,P,M\}$ | $M$ | $G$ |
| $D_2$ | $C_1$ | $G$ | $\{VP,P,M\}$ | $\{G,VG\}$ | $\{VP,P\}$ | $G$ |
| | $C_2$ | $\{G,VG\}$ | $\{P,M,G,VG\}$ | $\{VP,P,M\}$ | $P$ | $G$ |
| | $C_3$ | $M$ | $\{P,M\}$ | $M$ | $\{VP,P\}$ | $\{G,VG\}$ |
| | $C_4$ | $P$ | $M$ | $\{VP,P\}$ | $\{G,VG\}$ | $VG$ |
| $D_3$ | $C_1$ | $\{VP,P\}$ | $M$ | $\{G,VG\}$ | $\{VP,P,M\}$ | $P$ |
| | $C_2$ | $\{M,G,VG\}$ | $\{G,VG\}$ | $G$ | $M$ | $\{G,VG\}$ |
| | $C_3$ | $P$ | $\{M,G,VG\}$ | $\{P,M,G,VG\}$ | $VP$ | $G$ |
| | $C_4$ | $G$ | $\{VP,P\}$ | $\{G,VG\}$ | $\{VP,P\}$ | $VP$ |
| $D_4$ | $C_1$ | $VG$ | $\{G,VG\}$ | $\{VP,P\}$ | $M$ | $G$ |
| | $C_2$ | $\{P,M\}$ | $M$ | $VP$ | $\{G,VG\}$ | $VG$ |
| | $C_3$ | $\{VP,P\}$ | $\{G,VG\}$ | $M$ | $M$ | $\{VP,P\}$ |
| | $C_4$ | $M$ | $\{M,G\}$ | $\{VP,P,M\}$ | $P$ | $M$ |

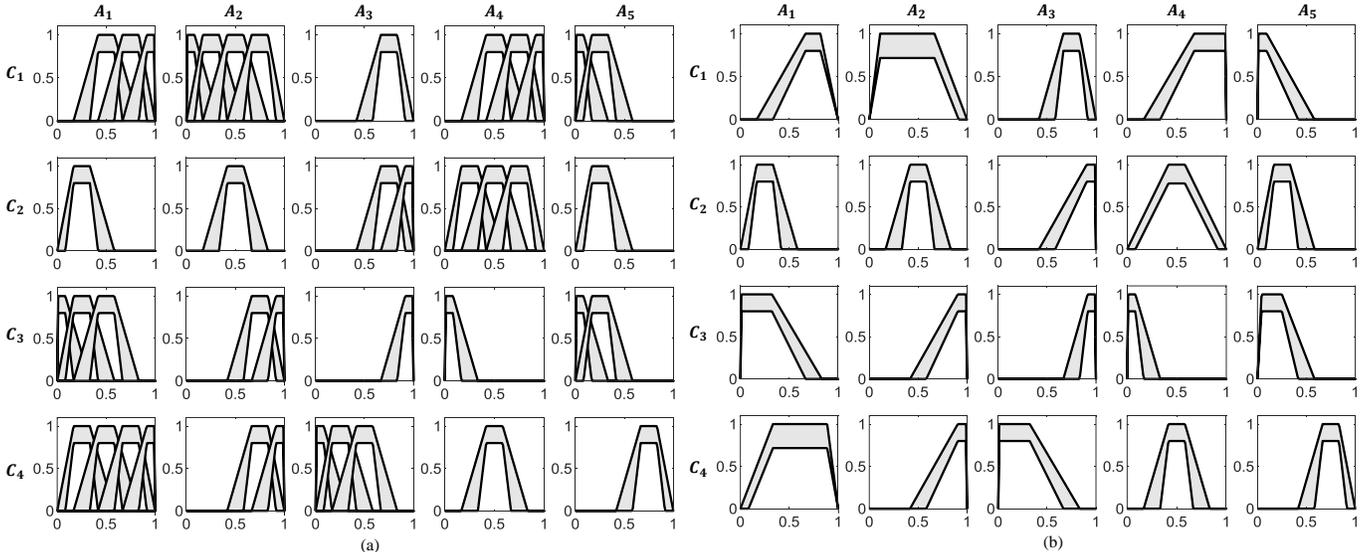

Fig. 11. Responses of DMR, $D_1$. (a) T2 HFLTSs corresponding to the responses in Table I, and (b) computed T2 fuzzy envelopes

3) *Envelope computation*

As an example, the T1 fuzzy envelopes of the $UMF$ and $LMF$ for $ll =$ between $M$ and $VG \equiv \widetilde{H}_{\widetilde{S}} = \{M,G,VG\}$ are

$$F_{\widetilde{H}_{\widetilde{S}}}^{UMF} \approx T_U(0.167, 0.667, 0.819, 1)$$

and

$$F_{\widetilde{H}_{\widetilde{S}}}^{LMF} \approx T_L(0.333, 0.667, 0.819, 1).$$

Then the entropies are computed using Eq. (9)-Eq. (13), Eq. (23) and Eq. (24) as follows :

$$\tilde{E}_f(\tilde{H}_{\tilde{S}}) = \frac{1}{3}\begin{pmatrix}\left(4 \times 0.040 \times \frac{2}{4} \times \left(1 - \frac{2}{4}\right)\right) + \\ \left(4 \times 0.051 \times \frac{3}{4} \times \left(1 - \frac{3}{4}\right)\right) + \\ \left(4 \times 0.040 \times \frac{4}{4} \times \left(1 - \frac{4}{4}\right)\right)\end{pmatrix} \approx 0.026$$

$$E_h(\tilde{H}_{\tilde{S}}) = \frac{1}{4} \times \frac{2}{3(3-1)} \times \sum_{i=1}^{2}\sum_{j=2}^{3}\left(I(\tilde{s}_{\alpha_j}) - I(\tilde{s}_{\alpha_i})\right) \approx 0.333$$

$$\beta(\tilde{H}_{\tilde{S}}) = \frac{1}{2}\left(\cos\left(\frac{\pi}{4} \times 2\right) + \sin\left(\left(\frac{\pi}{4} \times 4\right) - \frac{\pi}{2}\right)\right) = 0.5$$

$$\tilde{E}_c(\tilde{H}_{\tilde{S}}) = \frac{0.03 + (0.5 \times 0.33)}{1 + (0.5 \times 0.33)} \approx 0.165$$

i. The T2 fuzzy envelope is then computed using Eq. (25) and Eq. (26):

$$\bar{\mu}_{\tilde{F}_{\tilde{H}_{\tilde{S}}}}(x) = T_U(0.167, 0.667, 0.819, 1)$$

$$\underline{\mu}_{\tilde{F}_{\tilde{H}_{\tilde{S}}}}(x) \approx T_L(0.333, 0.667, 0.819, 1, 0.8)$$

The computed T2 fuzzy envelopes for $D_1$ are given in Fig 11(b). The same for the rest of the DMRs are given in Fig. SM-1 of the supplementary materials, SM-III.

4) *Aggregation of responses*

The obtained T2 fuzzy envelopes are then aggregated using Eq. (28). The result of aggregation for $D_1$ corresponding to supplier $A_1$ is given by the IT2 TrFN

$T\{(0, 0.180, 0.641, 0.975), (0.040, 0.173, 0.668, 0.893), 0.716\}$.

The aggregated responses for all the DMRs are shown in Fig 12.

5) *Ranking*

i. The ranking of suppliers obtained for all the DMRs on the basis of their centroids is given below:

$$D_1: A_2 \succ A_3 \succ A_5 \succ A_1 \succ A_4$$
$$D_2: A_5 \succ A_4 \succ A_2 \succ A_1 \succ A_3$$
$$D_3: A_3 \succ A_1 \succ A_2 \succ A_5 \succ A_4$$
$$D_4: A_2 \succ A_5 \succ A_4 \succ A_1 \succ A_3$$

This might also be seen as the rank matrix, $R_{4 \times 5}$.

ii. The scores for supplier $A_1$ are computed using Eq. (29) and Eq. (30). The calculations are given below.

    a. $\mathcal{O}(A_1) = \{(1,4), (2,4), (3,2), (4,4)\}$
    b. $Y = \{2, 4\}$
    c. $f(A_1, 2) = 1$
    d. $f(A_1, 4) = 3$

Therefore, using the above information
$$C(1,2) = 0.150$$
$$C(1,4) = \frac{1}{3}(0.250 + 0.400 + 0.200) \approx 0.283$$

$$\mathfrak{S}(A_1) = \frac{\left(((5+1) - 2) \times C(1,2)\right) + \left(((5+1) - 4) \times C(1,4)\right)}{C(1,2) + C(1,4)}$$

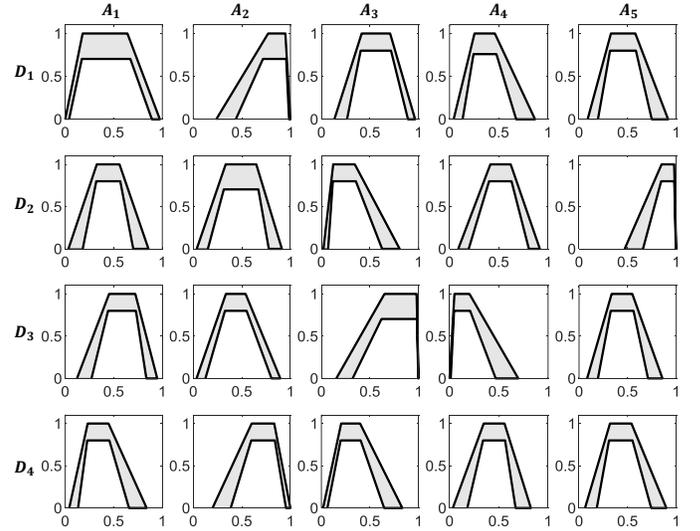

Fig. 12. Aggregated responses of every DMR.

$$\mathfrak{S}(A_1) = \frac{(4 \times 0.150) + (2 \times 0.283)}{0.283 + 0.150} \approx 2.692$$

Calculating similarly for rest of the suppliers, the score values for all of them are given below:

$\mathfrak{S} \approx 2.692, \mathfrak{S}(A_2) \approx 3.387, \mathfrak{S}(A_3) \approx 2.929, \mathfrak{S}(A_4) = 3.000, \mathfrak{S}(A_5) = 4.250$.

The final ranking of suppliers hence generated using the score values is:

$$A_5 \succ A_2 \succ A_4 \succ A_3 \succ A_1.$$

Therefore, the best supplier is $A_5$ which is ranked first according to the score function $\mathfrak{S}$.

## VI. COMPARATIVE ANALYSIS OF LDM USING T2 HFLTS

Our proposed linguistic MCGDM framework using T2 HFLTSs is the first such work where T2 FSs have been used for the semantics of linguistic terms within the HFLTS framework. This allows the consideration of both the intra-uncertainty as well as the inter-uncertainty faced by the DMRs while providing their assessments in the presence of other DMRs. This way, the obtained results are much more reliable and closer to the analogy of DMRs. Moreover, the proposed unique ranking methodology ensures that the ranking takes into consideration the expertise levels of the DMRs, while making sure that the importance of every rank does not get hidden within the weights of the DMRs.

In this section, we provide comparative studies with some existing models to demonstrate the advantages of LDM using T2 HFLTSs.

### A. *Comparative analysis with existing literature*

Compared to all the existing LDM models that use T2 FSs for the semantics, we observed that our proposal of using T2 FSs for information representation in T2 HFLTSs, is the first which handles the expert level intra-uncertainty, inter-uncertainty as well as hesitation, all considered together. Therefore, this is the first work which handles the above-mentioned points and works with CLEs to generate T2 representations of them. Also the LDM model presented in this paper for linguistic MCDM based on T2 FSs is the first one to be employed to handle representations of CLEs and T2 HFLTSs for the generation of final result.

TABLE III RESPONSES OF DMR IN EXAMPLE 2

| Criteria; Alternatives | $A_1$ | $A_2$ | $A_3$ | $A_4$ | $A_5$ |
|---|---|---|---|---|---|
| $C_1$ | $G$ | more than $G$ | less than $P$ | $M$ | $G$ |
| $C_2$ | between $P$ and $M$ | $M$ | $VP$ | more than $G$ | $VG$ |
| $C_3$ | less than $P$ | more than $M$ | $M$ | $M$ | less than $P$ |
| $C_4$ | $M$ | between $M$ and $G$ | between $VP$ and $M$ | $P$ | $M$ |

Considering the model[1] in [17], it makes use of T1 FSs as representation models for linguistic information in their system. Then the T2 fuzzy envelopes are computed for the CLEs given in the assessments. In comparison with our proposals in this paper, we make the following observations.

1. Our proposal of using T2 FSs as representation models for linguistic information in the CLEs in the form of T2 HFLTSs surpasses the model in [17] because the latter does not handle the expert level uncertainties, namely the inter-uncertainty and the intra-uncertainty. However, T2 FSs provide an obvious advantage over T1 FSs when dealing with such uncertainties due to their higher degrees of freedom.
2. Secondly, the utilization of T2 FSs for semantics of linguistic information produces a good amount of reduction in computations. This is because, when an expert provides his/her responses for an alternative, he/she may face hesitation while choosing a single term from the fixed set of linguistic terms presented to him/her. However, it is unlikely that he/she faces hesitancy while providing responses for every case. For e.g., consider the responses of the DMRs given in Table I. To solve this DM problem using the semantics provided in [17], T2 fuzzy envelopes should be computed for each of the responses, including the single terms. Since, they deal with representations based on T1 FSs (for linguistic terms), they compute the T2 fuzzy envelopes for a single term, say, $s_i$, using the CLE 'between $s_i$ and $s_i$'. This obviously leads to unnecessary computations within the system. However, our proposed representation model based on T2 FSs eliminates such unnecessary computations since envelopes are not computed for single terms. It is good to point out that our T2 FS based representation model achieves a reduction of almost ~43% in computations, with the responses given in Table I, when compared with the model given in [17].
3. Since envelopes are not computed for single terms in our proposed representation model, there is no information loss. This is because the information elicited from linguistic terms is the exact same as to what the experts respond with, hence the results obtained from such a model are much more reliable and robust.
4. In addition to using T2 FSs for representation of linguistic information, we adopt the usage of IT2 TrFNs in our proposal, unlike the previous models employing T1 TFNs. This allows for the DMRs to express their certainty over a certain interval of information. This is again beneficial because TFNs are special cases of TrFNs. Hence the semantics adopted in our model prove to be more flexible, hence producing better results.
5. The computation of the fuzzy entropy for HFLTSs to obtain T2 fuzzy envelopes in [17] is based only on the indices, hence ignoring the fuzziness inherent within the linguistic terms in the HFLTS. The usage of fuzzy entropies of each linguistic term present in a T2 HFLTS to determine the fuzziness present in it produces compact $FOU$s of the resultant T2 fuzzy envelopes in addition to ensuring that no information loss during the computations occurs. The compactness of the $FOU$s is also contributed to by the new way of computation of the corresponding $LMF$s.

Since there are no models in the existing literature that deal with T2 FSs as well as expert level uncertainties and hesitation, we compare our proposed LDM model with various other LDM models as described below. Note that for both the comparisons, all the criteria considered are benefit criteria.

1. Since our proposed model is capable of being used in both GDM and single expert MCDM settings, we first compare our model with the one presented in [36]. The authors claimed that their model, which was based on the technique for order preference by similarity to ideal solutions (TOPSIS) methodology provided effective and reliable results. Example 2 shows the results of comparison with this model.

**Example 2:** The problem used for comparison is given in Table III.
The rankings of alternatives obtained using the likelihood-based comparison approach given in [36] is given below:
$$A_1 = A_2 \succ A_4 \succ A_5 \succ A_3,$$
whereas, when the alternatives are ranked on the basis of our proposed method, we obtain the following ranking:
$$A_2 \succ A_5 \succ A_4 \succ A_1 \succ A_3.$$
The obtained T2 fuzzy envelopes for the problem and the related calculations are given in the supplementary materials in Section SM-IV.

**Analysis**
The results produced by the model given in [36] produces indistinguishable rankings for the given responses, i.e. the model is unable to choose the best alternative. However, our proposed model produced rankings that are unique. One can also notice the difference in the ranks obtained by the two models. This is because the technique for order preference by similarity to ideal solutions (TOPSIS) uses compromising mechanisms to obtain the final results. Moreover, in [36] approximate ideal solutions are calculated to make likelihood-based comparisons.

---

[1]The model given in [17] is not solved here for comparison purposes because it uses T1 TFNs for semantics of linguistic terms. Moreover, the authors in [17] and [28] assume symmetric partition of the linguistic term set to be aggregated, which does not apply for TrFNs. This hence, does not provide common grounds to compare the results obtained by the model in [17].

The method proposed in this paper, considers the exact T2 FS representations of the CLEs and computes their centroids to give the final result. Therefore, our method produces accurate results, which are close to the DMRs cognition while performing simpler computations.

2. Next, we compare our model with the one presented in [37]. This is a MCGDM model that works under the IT2 fuzzy environment. The comparison with this model is depicted through Example 3.

**Example 3:** We use the same set of responses given in Table I, to compare the two models. The rankings obtained from the method in [37] are given below:
$$A_5 \succ A_2 \succ A_3 \succ A_1 \succ A_4.$$
Refer to the supplementary materials, Section SM-V for the computations.

**Analysis**

It can be seen clearly that the rankings differ. The first reason for this is because the MCGDM in [37] does not handle criteria weights with higher order uncertainty. Whereas, our model handles weights which are represented using T2 FSs. Since criteria weights are decided upon by the experts themselves, the notion of word and expert-level uncertainties applies for criteria weights too. Hence, the results produced by our model are closer to the thinking of all experts. The second reason for the difference in the rankings is that, the ranking methodology utilized by [37] is based on various parameters of a T2 FS. However, in our model, the rankings are made based only on the centroids of the T2 FS. Since the centroid of a T2 FS is unique, our model produces more accurate results. Another advantage of our proposed model over the one presented in [37] is that the former can be used to generate and hence, observe the rankings obtained for different DMRs, which, one is unable to obtain with the latter model. In this way, our model has better transparency in the rankings.

## VII. Conclusions and Discussions

CLEs provide a good departure from fixed single terms that are used during assessments. This is because humans inherently specify their knowledge using natural language, and face hesitation while choosing single terms. Therefore, HFLTSs have been successful in providing methods for computations on CLEs, which represent richer linguistic information.

A person faces hesitation along with uncertainty in defining the semantics of a linguistic term. Therefore, the use of at least IT2 FSs as semantics for linguistic terms is appropriate. Another motivation for the usage of T2 FSs is that there are cases where a group of DMRs faces disagreements with each other. To respect the opinion of every DMR, the usage of T2 FSs or above has been recommended.

Hence, in this paper we have proposed the usage of T2 FSs to define the semantics of linguistic terms. This enriches the linguistic information elicited from the CLEs along with producing reliable results as already shown by the examples considered in this paper.

In addition to this, the following have been proposed in this article:

- The semantics of linguistic terms are defined using IT2 TrFNs, which provide greater flexibility of information representation as TFNs are special cases of TrFNs.
- A new method for computing T2 fuzzy envelopes from the resulting T2 HFLTSs has been introduced in this paper. It takes into account the fuzzy entropies of the individual linguistic terms in the T2 HFLTS.
- A novel linguistic MCGDM framework using T2 HFLTSs has also been proposed, the working of which has been explained using an SPE problem. The proposed framework computes the T2 fuzzy envelopes and then results in ranking of the alternatives.
- A unique expertise and priority weightage based ranking method is proposed, which provides unique scores to the alternatives.
- Our proposal of T2 HFLTSs reduces the computations required to obtain the T2 fuzzy envelope. This is because in existing models, final computations require all the responses to be represented equivalently, i.e., if T2 fuzzy envelopes are computed for CLEs, then all other responses must also be represented by T2 FSs. Therefore, on the basis of our proposed representation of terms, T2 fuzzy envelopes are not computed for single terms, hence reducing unnecessary computations. In our illustrative example in Section IV, we obtain a reduction of approximately 45% in computations as compared to existing methodologies.
- Highlighting all the aforementioned points we must mention that our proposed approach provides good ability to generate CLEs and hence T2 HFLTSs are more reliable. In this way, our model handles hesitation, inter-uncertainty as well as intra-uncertainty faced by the DMRs right from the response generation phase. To our knowledge, no other existing T2 FS based LDM model so far has been able to handle all these uncertainties all taken together.

In the future, we shall consider applying our proposed model to many crucial application domains in real life scenarios. We shall also work with higher order uncertainty both in the information representation as well as in the computation phase.

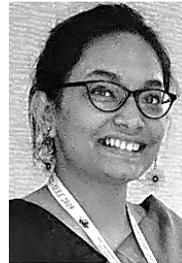

**Taniya Seth** (S'17) was born in New Delhi, India. She received her M.Sc. degree in computer science from South Asian University, New Delhi, India, in 2017. She is currently pursuing Ph.D. from South Asian University, New Delhi, India. Her research interests include, computing with words, decision making, intelligent systems, fuzzy logic, optimization and deep learning. She received the Gold Medal in her M.Sc. degree in 2017. She also received the Google WomenTechmaker Scholarship by Google in 2018. She is involved with many women in technology initiatives with organizations such as Google and Facebook. She has also published papers in well-known conferences such as IEEE-FUZZ and IEEE-WCCI. Currently, she is the recipient of DST INSPIRE Fellowship awarded by Department of Science and Technology, Ministry of Science and Technology, Government of India.

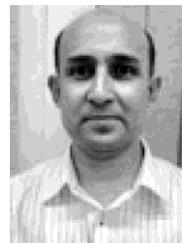

**Pranab K. Muhuri** (M'08, SM'19) was born in Chittagong, Bangladesh. He received his Ph.D. degree in Computer Engineering in 2005 from IT-BHU [now Indian Institute of Technology (BHU)], Varanasi, India. He is currently a Professor with the Department of Computer Science, South Asian University, New Delhi, India, where he is leading the computational intelligence research group. His current research interests are mainly in real-time systems, computational intelligence, especially fuzzy systems, evolutionary algorithms, perceptual computing, and machine learning. Pranab is an active member of the IEEE Computer Society, IEEE Computational Intelligence Society, IEEE SMC Society and Association of Computing Machinery (ACM). Pranab has published about 60 research papers in well-known journals and conferences including IEEE Transactions on Fuzzy Systems, Fuzzy Sets and Systems, Applied Soft Computing, Future Generation Computing Systems, Computers and Industrial Engineering, and Engineering Applications of Artificial Intelligence. Currently, he is serving as a member of the Editorial Board of the Applied Soft Computing journal.

# SUPPLEMENTARY MATERIALS

## SM-I

### PROOF OF FUZZY ENTROPY MEASURE FOR T2 HFLTS

Referring to Definition 7 and 8 in Section II and Definition 14 in Section IV, it is sufficient to prove that the proposed fuzzy entropy measure for T2 HFLTSs in Definition 15 satisfies the conditions (FE1)-(FE4).

Let $f\left(\frac{I(\tilde{s}_{\alpha_i})}{g}\right) = 4\left(\tilde{F}_{\tilde{s}_{\alpha_i}}\right) \times \frac{I(\tilde{s}_{\alpha_i})}{g}\left(1 - \frac{I(\tilde{s}_{\alpha_i})}{g}\right) \in [0,1]$.

- (FE1) $\tilde{E}_f(\tilde{H}_{\tilde{S}}) = 0$ when $I(\tilde{s}_{\alpha_i}) = 0$ or $I(\tilde{s}_{\alpha_i}) = g$, i.e. when $\tilde{H}_{\tilde{S}} = \{\tilde{s}_0\}$ or $\tilde{H}_{\tilde{S}} = \{\tilde{s}_g\}$.

- (FE2) $\tilde{E}_f\left(\left\{\tilde{s}_{\frac{g}{2}}\right\}\right)$ is a unique maximum of $\tilde{E}_f$ because, for any value of $\tilde{F}_{\tilde{s}_{\frac{g}{2}}} \in [0,1]$, $f\left(\frac{I\left(\tilde{s}_{\frac{g}{2}}\right)}{g}\right) = 1$.

- (FE3) If $\tilde{s}_{\alpha_i} \leq \tilde{s}_{\alpha_j} \leq \tilde{s}_{\frac{g}{2}}$, then $i \leq j$ and $\left(f\left(\frac{I(\tilde{s}_i)}{g}\right)\right) \leq \left(f\left(\frac{I(\tilde{s}_j)}{g}\right)\right)$. Thus for any value of $\tilde{F}_{\tilde{s}_{\alpha_i}}, \tilde{F}_{\tilde{s}_{\alpha_j}} \in [0,1]$, (FE3) holds.

- (FE4) It is easy to verify that $f\left(\frac{I(\tilde{s}_{\alpha_i})}{g}\right) = f\left(1 - \frac{I(\tilde{s}_{\alpha_i})}{g}\right)$ for any $i \in [0, g]$. (FE4) holds for LTS with symmetrical IT2 TrFNs.

## SM-II

### EKM ALGORITHMS FOR CENTROID CALCULATION

Let $\tilde{A}$ be an IT2 FS, defined on for the set $X$. Let $\boldsymbol{\theta}$ be the set of weights associated with every element of $X$. The centroid of $\tilde{A}$ is computed as [35]:

$$\frac{c_l + c_r}{2}$$

where,

$$c_l = \min_{\forall \theta_i \in [\underline{\mu}_{\tilde{A}}(x_i), \mu_{\tilde{A}}(x_i)]} \left(\frac{\sum_i^N x_i \theta_i}{\sum_i^N \theta_i}\right), x_i \in X, \theta_i \in \boldsymbol{\theta},$$

$$c_r = \max_{\forall \theta_i \in [\underline{\mu}_{\tilde{A}}(x_i), \mu_{\tilde{A}}(x_i)]} \left(\frac{\sum_i^N x_i \theta_i}{\sum_i^N \theta_i}\right), x_i \in X, \theta_i \in \boldsymbol{\theta}.$$

## SM-III

### EXAMPLE 1: COMPUTED T2 FUZZY ENVELOPES

The T2 HFLTSs and corresponding T2 fuzzy envelopes for the rest of the DMRs are given in Fig. SM-1. Fig. SM-1(a), Fig. SM-1(c), and Fig. SM-1(e) show the T2 HFLTSs for DMRs $D_1$, $D_2$, and $D_3$ respectively. Whereas, Fig. SM-1(b), Fig. SM-1(d), and Fig. SM-1(f) show the corresponding T2 fuzzy envelopes for DMRs $D_1$, $D_2$, and $D_3$ respectively.

## SM-IV

### EXAMPLE 2: MCDM WITH IT2 FS BASED TOPSIS USING T2 FUZZY ENVELOPES

For ease of understanding, we represent the T2 FSs as given in [36]. We proceed step-wise as given in the paper.

Step 1: The MCDM problem is formulated with alternative set $A = \{A_1, A_2, A_3, A_4, A_5\}$ and criteria set $C = \{C_1 C_2, C_3, C_4\}$, all of which are benefit criteria.

Step 2: The obtained CLEs are converted to T2 HFLTSs and corresponding T2 fuzzy envelopes are computed. They are given in Fig. SM-2.

Step 3: The evaluative ratings $E_{ij}$ are the T2 fuzzy envelopes computed for alternative $A_i \in A$ and criterion $c_j \in C$, as shown in Fig. SM-2.

Step 4: With evaluative ratings $E_{ij}$ and criteria weights $W_j$, represented by IT2 TrFNs, they are defined as follows.

$$E_{ij} = [E_{ij}^-, E_{ij}^+] =$$
$$\left[\left(a_{1ij}^-, a_{2ij}^-, a_{3ji}^-, a_{4ij}^-; h_{E_{ij}}^-\right), \left(a_{1ij}^+, a_{2ij}^+, a_{3ij}^+, a_{4ij}^+; h_{E_{ij}}^+\right)\right],$$
$$W_j = [W_j^-, W_j^+] =$$
$$\left[\left(w_{1j}^-, w_{2j}^-, w_{3j}^-, w_{4j}^-; h_{w_j}^-\right), \left(w_{1j}^+, w_{2j}^+, w_{3j}^+, w_{4j}^+; h_{w_j}^+\right)\right].$$

The weighted evaluative rating is obtained by:

$$\overline{E}_{ij} = W_j \otimes E_{ij} = \left[\left(w_{1j}^- \cdot \overline{a}_{1ij}^-, w_{2j}^- \cdot \overline{a}_{2ij}^-, w_{3j}^- \cdot \overline{a}_{3ij}^-, w_{4j}^- \cdot \overline{a}_{4ij}^-; \min\{h_{w_j}^-, h_{E_{ij}}^-\}\right), \left(w_{1j}^+ \cdot \overline{a}_{1ij}^+, w_{2j}^+ \cdot \overline{a}_{2ij}^+, w_{3j}^+ \cdot \overline{a}_{3ij}^+, w_{4j}^+ \cdot \overline{a}_{4ij}^+; \min\{h_{w_j}^+, h_{E_{ij}}^+\}\right)\right].$$

Step 5 and 6: The weighted evaluative ratings $\overline{E}_{\rho j}$ and $\overline{E}_{\eta j}$ of the approximate positive-ideal and negative-ideal solutions $A_\rho$ and $A_\eta$, respectively with respect to criterion $c_j \in C$ are computed as follows

$$\overline{E}_{\rho j} = \begin{bmatrix} \left(\vee_{i=1}^5 \overline{a}_{1ij}^-, \vee_{i=1}^5 \overline{a}_{2ij}^-, \vee_{i=1}^5 \overline{a}_{3ji}^-, \vee_{i=1}^5 \overline{a}_{4ij}^-; \wedge_{i=1}^5 h_{\overline{E}_{ij}}^-\right), \\ \left(\vee_{i=1}^5 \overline{a}_{1ij}^+, \vee_{i=1}^5 \overline{a}_{2ij}^+, \vee_{i=1}^5 \overline{a}_{3ji}^+, \vee_{i=1}^5 \overline{a}_{4ij}^+; \wedge_{i=1}^5 h_{\overline{E}_{ij}}^+\right) \end{bmatrix}$$

$$\overline{E}_{\eta j} = \begin{bmatrix} \left(\wedge_{i=1}^5 \overline{a}_{1ij}^-, \wedge_{i=1}^5 \overline{a}_{2ij}^-, \wedge_{i=1}^5 \overline{a}_{3ji}^-, \wedge_{i=1}^5 \overline{a}_{4ij}^-; \wedge_{i=1}^5 h_{\overline{E}_{ij}}^-\right), \\ \left(\wedge_{i=1}^5 \overline{a}_{1ij}^+, \wedge_{i=1}^5 \overline{a}_{2ij}^+, \wedge_{i=1}^5 \overline{a}_{3ji}^+, \wedge_{i=1}^5 \overline{a}_{4ij}^+; \wedge_{i=1}^5 h_{\overline{E}_{ij}}^+\right) \end{bmatrix}$$

It is easy to obtain the approximate positive-ideal and negative-ideal solutions as follows:

$\overline{E}_{\rho 1} = [(0,0,0.083,0.250; 0.8), (0,0,0.083,0.417; 1)]$

$\overline{E}_{\rho 2} = [(0.069,0.229,0.417,0.583; 0.8), (0,0.229,0.417,0.750; 1)]$

$\overline{E}_{\rho 3} = [(0.243,0.528,0.750,0.915; 0.8), (0.104,0.528,0.750,1; 1)]$

$\overline{E}_{\rho 4} = [(0.250,0.457,0.750,0.915; 0.8), (0.097,0.457,0.750,1; 1)]$

$\overline{E}_{\eta 1} = [(0,0,0.008,0.104; 0.8), (0,0,0.008,0.243; 1)]$

$\overline{E}_{\eta 2} = [(0,0,0.035,0.097; 0.8), (0,0,0.035,0.250; 1)]$

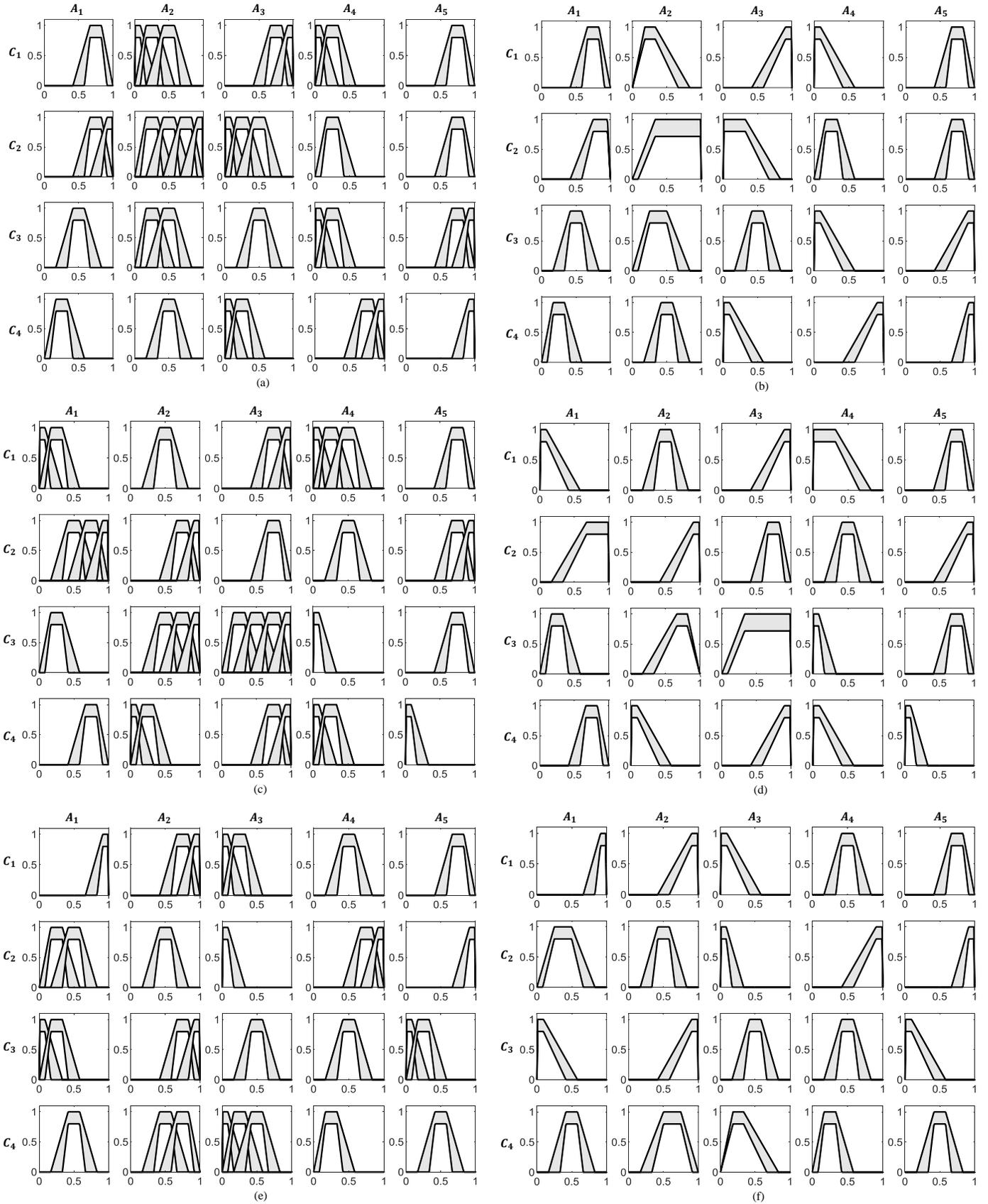

Fig. SM- 1. T2 HFLTSs and corresponding T2 fuzzy envelopes of DMRs

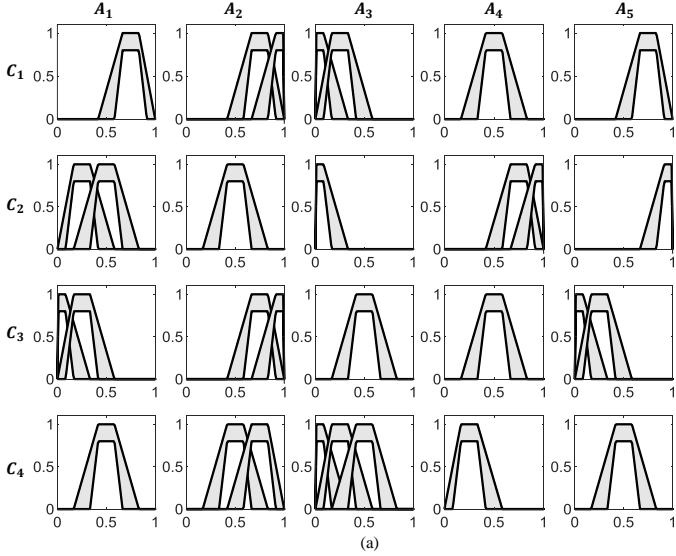
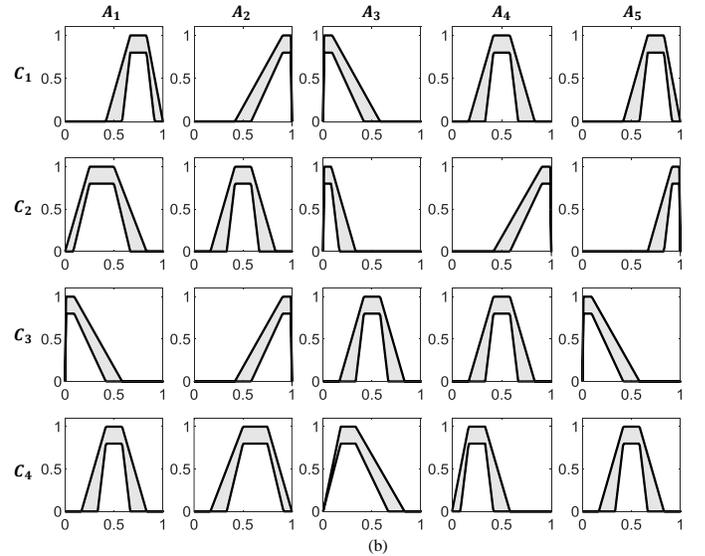

Fig. SM- 2. T2 HFLTSs and T2 fuzzy envelopes of DMR in Example 2

$\overline{E}_{\eta 3} = [(0,0,0.070,0.381; 0.8), (0,0,0.070,0.583; 1)]$
$\overline{E}_{\eta 4} = [(0,0.152,0.333,0.417; 0.8), (0,0.152,0.333,0.583; 1)]$

Step 7 and 8: Let $X$ and $Y$ be defined as follows:
$$X_{ij} = [X_{ij}^-, X_{ij}^+] = \left[\left(x_{1ij}^-, x_{2ij}^-, x_{3ji}^-, x_{4ij}^-; h_{X_{ij}}^-\right), \left(x_{1ij}^+, x_{2ij}^+, x_{3ij}^+, x_{4ij}^+; h_{X_{ij}}^+\right)\right],$$
$$Y_{ij} = [Y_{ij}^-, Y_{ij}^+] = \left[\left(y_{1ij}^-, y_{2ij}^-, y_{3ji}^-, y_{4ij}^-; h_{Y_{ij}}^-\right), \left(y_{1ij}^+, y_{2ij}^+, y_{3ij}^+, y_{4ij}^+; h_{Y_{ij}}^+\right)\right].$$

Let $\varsigma$ be any positive integer. Assuming that at least one of $h_X^- \neq h_Y^+$, $x_4^- \neq x_1^+$, $y_4^+ \neq y_1^+$, $x_\varsigma^- \neq y_\varsigma^+$ and at least one of $h_X^+ \neq h_Y^-$, $x_4^+ \neq x_1^+$, $y_4^- \neq y_1^-$, $x_\varsigma^+ \neq y_\varsigma^-$ hold, where $\varsigma = 1,2,3,4$, the lower and upper likelihood-based comparison indices for the relation $\geq$ between sets $X$ and $Y$ is computed as follows:

$$LI^-(X \geq Y) = \max\{1 - \mathfrak{M}, 0\},$$

$$\mathfrak{M} = \max \left[ \frac{\sum_{\varsigma=1}^4 \max(y_\varsigma^+ - x_\varsigma^-, 0) + (y_4^+ - x_1^-) + 2\max(h_Y^+ - h_X^-, 0)}{\sum_{\varsigma=1}^4 |y_\varsigma^+ - x_\varsigma^-| + (x_4^- - x_1^-) + (y_4^+ - y_1^+) + 2|h_Y^+ - h_X^-|}, 0 \right]$$

$$LI^+(X \geq Y) = \max\{1 - \mathfrak{N}, 0\},$$

$$\mathfrak{N} = \max \left[ \frac{\sum_{\varsigma=1}^4 \max(y_\varsigma^- - x_\varsigma^+, 0) + (y_4^- - x_1^+) + 2\max(h_Y^- - h_X^+, 0)}{\sum_{\varsigma=1}^4 |y_\varsigma^- - x_\varsigma^+| + (x_4^+ - x_1^+) + (y_4^- - y_1^-) + 2|h_Y^- - h_X^+|}, 0 \right]$$

The likelihood-based comparison index $LI(X \geq Y)$ is computed as
$$LI(X \geq Y) = \frac{LI^-(X \geq Y) + LI^+(X \geq Y)}{2}$$

The values of comparison indices of the weighted evaluative ratings with respect to the approximate positive-ideal and negative-ideal solutions are given below. They are given in the form of a matrix.

$$LI(\overline{E}_{\rho j} \geq \overline{E}_{ij}) = \begin{bmatrix} 0.597 & 0.807 & 0.937 & 0.941 \\ 0.643 & 0.830 & 0.890 & 0.927 \\ 0.753 & 0.921 & 0.925 & 0.950 \\ 0.723 & 0.469 & 0.932 & 0.970 \\ 0.720 & 0.852 & 0.959 & 0.957 \end{bmatrix},$$

$$LI(\overline{E}_{ij} \geq \overline{E}_{\eta j}) = \begin{bmatrix} 0.987 & 0.988 & 0.977 & 0.987 \\ 0.987 & 0.988 & 0.986 & 0.987 \\ 0.985 & 0.985 & 0.984 & 0.983 \\ 0.986 & 0.989 & 0.985 & 0.982 \\ 0.987 & 0.990 & 0.979 & 0.988 \end{bmatrix}.$$

Step 9: The likelihood-based closeness coefficient for every alternative is computed using the following formulation:
$$LC_i = \frac{\sum_{j=1}^4 LI(\overline{E}_{ij} \geq \overline{E}_{\eta j})}{\sum_{j=1}^4 \left(LI(\overline{E}_{\rho j} \geq \overline{E}_{ij}) + LI(\overline{E}_{ij} \geq \overline{E}_{\eta j})\right)}$$

The final computed value for every alternative in our considered example is:
$$LC_1 \approx 0.546, LC_2 \approx 0.546, LC_3 \approx 0.526, LC_4 \approx 0.532,$$
$$LC_5 \approx 0.531$$

Step 10: The ranking process therefore gives the ranking $A_1 = A_2 \succ A_4 \succ A_5 \succ A_3$.

## SM-V

### EXAMPLE 3: MCGDM WITH IT2 FS BASED METHOD USING T2 FUZZY ENVELOPES

Here, we give the step-wise computations done while solving the MCGDM problem, given in Table I, in Section V of the article. using the model given in [37]. The corresponding T2 HFLTSs and T2 fuzzy envelopes are given in Fig. 11 and Fig. SM-1. Again for the convenience of comparison, the representation of IT2 TrFNs is adopted from [37]. Therefore, in this Section, an IT2 TrFN $\tilde{E}_i$ defined on $X$ is represented as given below. The following parameters are given in Fig. SM-3.

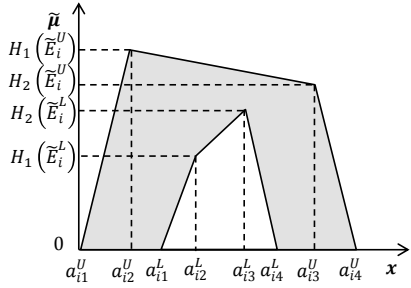

Fig. SM-3. An IT2 TrFN as given in [37]

$$\tilde{E}_i = \begin{pmatrix} \left(a_{i1}^U, a_{i2}^U, a_{i3}^U, a_{i4}^U: H_1(\tilde{E}_i^U), H_2(\tilde{E}_i^U)\right), \\ \left(a_{i1}^L, a_{i2}^L, a_{i3}^L, a_{i4}^L: H_1(\tilde{E}_i^L), H_2(\tilde{E}_i^L)\right) \end{pmatrix}$$

Step 1: Computed T2 fuzzy envelopes of responses are denoted as IT2 fuzzy decision matrices $\tilde{R}^k = (\tilde{E}_{ij}^k)_{n \times m}$ where $k = 1,2,3,4$ for $k$ DMRs in the system, $n = 1,2,3,4,5$ attributes and $m = 1,2,3,4$ criteria, with $\lambda = (\lambda_1, \lambda_2, \lambda_3, \lambda_4)^T$ and $\omega = (\omega_1, \omega_2, \omega_3, \omega_4)^T$ as the weights for DMRs and criteria respectively such that $\sum_{i=1}^{4} \lambda_i = 1$, $\sum_{k=1}^{4} \omega_k = 1$. Obtain collective normalized IT2 fuzzy decision matrix $\tilde{R} = (\tilde{E}_{ij})_{n \times m}$ as follows:

$$\tilde{E}_{ij} = \oplus_{k=1}^{4} \lambda_k \tilde{E}_{ij}^k$$

where, the operation $\oplus$ between two IT2 TrFNs is defined as

$$\tilde{E}_1 \oplus \tilde{E}_2 = \begin{pmatrix} \begin{pmatrix} a_{11}^U + a_{21}^U, a_{12}^U + a_{22}^U, a_{13}^U + a_{23}^U, a_{14}^U + a_{24}^U \\ :\min\{H_1(\tilde{E}_1^U), H_1(\tilde{E}_2^U)\}, \min\{H_2(\tilde{E}_1^U), H_2(\tilde{E}_2^U)\} \end{pmatrix}, \\ \begin{pmatrix} a_{11}^L + a_{21}^L, a_{12}^L + a_{22}^L, a_{13}^L + a_{23}^L, a_{14}^L + a_{24}^L \\ :\min\{H_1(\tilde{E}_1^L), H_1(\tilde{E}_2^L)\}, \min\{H_2(\tilde{E}_1^L), H_2(\tilde{E}_2^L)\} \end{pmatrix} \end{pmatrix}.$$

For our problem $\tilde{R}$ is computed as given below. The matrix is given column wise for easy representation.

$$\tilde{R}_1 = \begin{bmatrix} \begin{pmatrix} (0.342, 0.616, 0.752, 0.938: 1,1), \\ (0.4832, 0.616, 0.752, 0.878: 0.8, 0.8) \end{pmatrix} \\ \begin{pmatrix} (0.108, 0.343, 0.587, 0.908: 1,1), \\ (0.167, 0.343, 0.587, 0.795: 0.716, 0.716) \end{pmatrix} \\ \begin{pmatrix} (0.333, 0.665, 0.777, 0.917: 1,1), \\ (0.467, 0.665, 0.777, 0.862: 0.8, 0.8) \end{pmatrix} \\ \begin{pmatrix} (0.075, 0.253, 0.453, 0.775: 1,1), \\ (0.150, 0.253, 0.453, 0.650: 0.8, 0.8) \end{pmatrix} \\ \begin{pmatrix} (0.313, 0.5, 0.648, 0.896: 1,1), \\ (0.437, 0.5, 0.648, 0.790: 0.8, 0.8) \end{pmatrix} \end{bmatrix}$$

$$\tilde{R}_2 = \begin{bmatrix} \begin{pmatrix} (0.192, 0.493, 0.716, 0.862: 1,1), \\ (0.321, 0.493, 0.716, 0.79: 0.8, 0.8) \end{pmatrix} \\ \begin{pmatrix} (0.138, 0.456, 0.812, 0.925: 1,1), \\ (0.271, 0.456, 0.812, 0.850: 0.716, 0.716) \end{pmatrix} \\ \begin{pmatrix} (0.167, 0.326, 0.521, 0.8: 1,1), \\ (0.233, 0.326, 0.521, 0.687: 0.8, 0.8) \end{pmatrix} \\ \begin{pmatrix} (0.108, 0.414, 0.567, 0.808: 1,1), \\ (0.221, 0.414, 0.567, 0.695: 0.778, 0.778) \end{pmatrix} \\ \begin{pmatrix} (0.362, 0.627, 0.767, 0.896: 1,1), \\ (0.508, 0.627, 0.767, 0.820: 0.8, 0.8) \end{pmatrix} \end{bmatrix}$$

$$\tilde{R}_3 = \begin{bmatrix} \begin{pmatrix} (0.067, 0.192, 0.383, 0.746: 1,1), \\ (0.146, 0.192, 0.383, 0.580: 0.8, 0.8) \end{pmatrix} \\ \begin{pmatrix} (0.213, 0.607, 0.773, 0.933: 1,1), \\ (0.346, 0.607, 0.773, 0.867: 0.8, 0.8) \end{pmatrix} \\ \begin{pmatrix} (0.267, 0.528, 0.750, 0.9: 1,1), \\ (0.421, 0.528, 0.750, 0.8: 0.716, 0.716) \end{pmatrix} \\ \begin{pmatrix} (0.033, 0.083, 0.187, 0.533: 1,1), \\ (0.067, 0.083, 0.187, 0.367: 0.8, 0.8) \end{pmatrix} \\ \begin{pmatrix} (0.230, 0.473, 0.606, 0.812: 1,1), \\ (0.321, 0.472, 0.606, 0.725: 0.8, 0.8) \end{pmatrix} \end{bmatrix}$$

$$\tilde{R}_4 = \begin{bmatrix} \begin{pmatrix} (0.096, 0.333, 0.6, 0.8: 1,1), \\ (0.208, 0.333, 0.6, 0.687: 0.716, 0.716) \end{pmatrix} \\ \begin{pmatrix} (0.204, 0.493, 0.647, 0.871: 1,1), \\ (0.346, 0.493, 0.647, 0.762: 0.8, 0.8) \end{pmatrix} \\ \begin{pmatrix} (0.063, 0.172, 0.335, 0.758: 1,1), \\ (0.088, 0.172, 0.335, 0.617: 0.8, 0.8) \end{pmatrix} \\ \begin{pmatrix} (0.208, 0.5, 0.627, 0.812: 1,1), \\ (0.033, 0.5, 0.627, 0.712: 0.8, 0.8) \end{pmatrix} \\ \begin{pmatrix} (0.404, 0.616, 0.737, 0.867: 1,1), \\ (0.546, 0.616, 0.737, 0.787: 0.8, 0.8) \end{pmatrix} \end{bmatrix}$$

Step 2: Get result of weighted arithmetic average as follows:

$$\tilde{E}_i = \oplus_{j=1}^{4} \omega_j \tilde{E}_{ij},$$

$$\tilde{E}_1 = \begin{pmatrix} (0.131, 0.351, 0.572, 0.810: 1,1), \\ (0.240, 0.351, 0.572, 0.640: 0.767, 0.767) \end{pmatrix}$$

$$\tilde{E}_2 = \begin{pmatrix} (0.184, 0.505, 0.712, 0.904: 1,1), \\ (0.313, 0.505, 0.712, 0.814: 0.775, 0.775) \end{pmatrix}$$

$$\tilde{E}_3 = \begin{pmatrix} (0.172, 0.360, 0.541, 0.825: 1,1), \\ (0.255, 0.360, 0.541, 0.710: 0.775, 0.775) \end{pmatrix}$$

$$\tilde{E}_4 = \begin{pmatrix} (0.123, 0.333, 0.465, 0.724: 1,1), \\ (0.213, 0.333, 0.465, 0.6: 0.796, 0.796) \end{pmatrix}$$

$$\tilde{E}_4 = \begin{pmatrix} (0.334, 0.564, 0.695, 0.859: 1,1), \\ (0.460, 0.564, 0.695, 0.775: 0.8, 0.8) \end{pmatrix}$$

Step 3: The overall values obtained for the alternatives above are used to compute the rank values. For any IT2 TrFN $\tilde{E}_i$, its rank value, $Rank(\tilde{E}_i)$ is computed as follows:

$$\begin{aligned} Rank(\tilde{E}_i) &= M_1(\tilde{E}_i^U) + M_1(\tilde{E}_i^L) + M_2(\tilde{E}_i^U) + M_2(\tilde{E}_i^L) \\ &\quad + M_3(\tilde{E}_i^U) + M_4(\tilde{E}_i^L) \\ &\quad - \frac{1}{4}\Big(S_1(\tilde{E}_i^U) + S_1(\tilde{E}_i^L) + S_2(\tilde{E}_i^U) + S_2(\tilde{E}_i^L) \\ &\quad + S_3(\tilde{E}_i^U) + S_3(\tilde{E}_i^L) + S_4(\tilde{E}_i^U) + S_4(\tilde{E}_i^L)\Big) \\ &\quad + H_1(\tilde{E}_i^U) + H_1(\tilde{E}_i^L) + H_2(\tilde{E}_i^U) + H_2(\tilde{E}_i^L) \end{aligned}$$

where, $M_p(\tilde{E}_i^j) = \frac{a_{ip}^j + a_{i(p+1)}^j}{2}$, $S_p(\tilde{E}_i^j) = \sqrt{\frac{1}{2}\sum_{k=p}^{p+1}\left(a_{ik}^j - M_p(\tilde{E}_i^j)\right)^2}$ for $1 \le p \le 3$, $S_4(\tilde{E}_i^j) = \sqrt{\frac{1}{2}\sum_{k=1}^{4}\left(a_{ik}^j - \frac{1}{4}\sum_{k=1}^{4} a_{ik}^j\right)^2}$, $j \in \{U, L\}$.

The ranking values for our problem are:
$Rank(\tilde{E}_1) = 6.067$, $Rank(\tilde{E}_2) = 6.823$, $Rank(\tilde{E}_3) = 6.088$, $Rank(\tilde{E}_4) = 5.803$, and $Rank(\tilde{E}_5) = 7.148$.

Step 4: The rankings hence obtained for each alternative are $A_5 \succ A_2 \succ A_3 \succ A_1 \succ A_4$.